%% file: main.tex
\documentclass[10pt,twocolumn,letterpaper]{article}

\usepackage[pagenumbers]{cvpr} 

\input{preamble}

\definecolor{cvprblue}{rgb}{0.21,0.49,0.74}
\usepackage[pagebackref,breaklinks,colorlinks]{hyperref}

\usepackage[capitalize]{cleveref}
\crefname{section}{Sec.}{Secs.}
\Crefname{section}{Section}{Sections}
\Crefname{table}{Table}{Tables}
\crefname{table}{Tab.}{Tabs.}

\def\paperID{*****} 
\def\confName{CVPR}
\def\confYear{2024}

\title{Reference-based Restoration of Digitized Analog Videotapes}

\author{Lorenzo Agnolucci \and Leonardo Galteri \and  Marco Bertini \and Alberto Del Bimbo \vspace{0.1ex} \and
University of Florence - Media Integration and Communication Center (MICC) \\
Florence, Italy\\
{\tt\small [name.surname]@unifi.it}
}

\begin{document}
\maketitle
\input{sec/0_abstract}    
\input{sec/1_intro}
\input{sec/2_related}
\input{sec/3_method}
\input{sec/4_results}
\input{sec/5_conclusion}
{
    \small
    \bibliographystyle{ieeenat_fullname}
    \bibliography{main}
}

\input{sec/X_suppl}

\end{document}

%% file: preamble.tex
\usepackage[dvipsnames]{xcolor}

\usepackage{graphicx,scalerel}
\usepackage{amsmath}
\usepackage{amssymb}
\usepackage{booktabs}

\newcommand{\hide}[1]{}

\usepackage{color}
\usepackage{xcolor,colortbl}
\usepackage{mathtools}
\usepackage{multirow}
\usepackage{bm} 
\definecolor{purple}{rgb}{0.65,0,0.65}
\definecolor{dark_green}{rgb}{0, 0.5, 0}
\definecolor{blueish}{rgb}{0.0, 0.3, .6}
\definecolor{tabhighlight}{HTML}{e5e5e5}
\newcolumntype{h}{>{\columncolor{tabhighlight}}c}

\makeatletter
\newcommand*\wt[2][0.15ex]{%
        \begingroup
        \mathchoice{\wt@helper{#1}{#2}{\displaystyle}{\textfont}}
                   {\wt@helper{#1}{#2}{\textstyle}{\textfont}}
                   {\wt@helper{#1}{#2}{\scriptstyle}{\scriptfont}}
                   {\wt@helper{#1}{#2}{\scriptscriptstyle}{\scriptscriptfont}}%
        \endgroup
        #2%
}
\newcommand*\wt@helper[4]{%
        \def\currentfont{\the#41}%
        \def\currentskewchar{\char\the\skewchar\currentfont}%
        \setbox\tw@\hbox{\currentfont#2\currentskewchar}%
        \dimen@ii\wd\tw@
        \setbox\tw@\hbox{\currentfont#2{}\currentskewchar}%
        \advance\dimen@ii-\wd\tw@
        \rlap{\raisebox{-#1}{$\m@th#3\kern\dimen@ii\widetilde{\phantom{#2}}$}}%
}
\makeatother

\usepackage[export]{adjustbox} 

\newlength{\gridimagewidth}
\usepackage{array}
\newcolumntype{C}[1]{>{\centering\let\newline\\\arraybackslash\hspace{0pt}}m{#1}}

\newcolumntype{h}{>{\columncolor{tabhighlight}}c}

\newlength{\syntheticwidth}
\newlength{\realworldwidth}
\newcommand{\method}{TAPE\xspace}
\newcommand{\methodextended}{resToration of digitized Analog videotaPEs\xspace}

%% file: sec/0_abstract.tex
\begin{abstract}
Analog magnetic tapes have been the main video data storage device for several decades. Videos stored on analog videotapes exhibit unique degradation patterns caused by tape aging and reader device malfunctioning that are different from those observed in film and digital video restoration tasks. In this work, we present a reference-based approach for the \methodextended (\method). We leverage CLIP for zero-shot artifact detection to identify the cleanest frames of each video through textual prompts describing different artifacts. Then, we select the clean frames most similar to the input ones and employ them as references. We design a transformer-based Swin-UNet network that exploits both neighboring and reference frames via our Multi-Reference Spatial Feature Fusion (MRSFF) blocks. MRSFF blocks rely on cross-attention and attention pooling to take advantage of the most useful parts of each reference frame. To address the absence of ground truth in real-world videos, we create a synthetic dataset of videos exhibiting artifacts that closely resemble those commonly found in analog videotapes. Both quantitative and qualitative experiments show the effectiveness of our approach compared to other state-of-the-art methods. The code, the model, and the synthetic dataset are publicly available at \small{\href{https://github.com/miccunifi/TAPE}{\url{https://github.com/miccunifi/TAPE}}}.
\end{abstract}

%% file: sec/1_intro.tex
\section{Introduction}
\label{sec:intro}

\begin{figure}
  \centering
  \begin{subfigure}{0.32625\linewidth}
    \includegraphics[width=\linewidth]{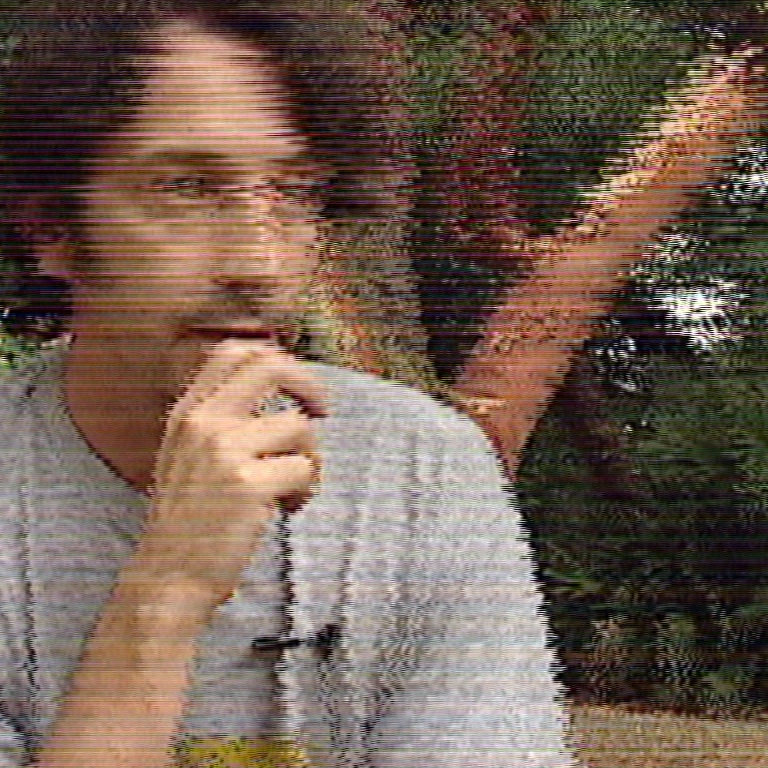}
    \caption{Frame 0}
    \label{fig:introduction_frame_a}
  \end{subfigure}
  \hfill
  \begin{subfigure}{0.32625\linewidth}
    \includegraphics[width=\linewidth]{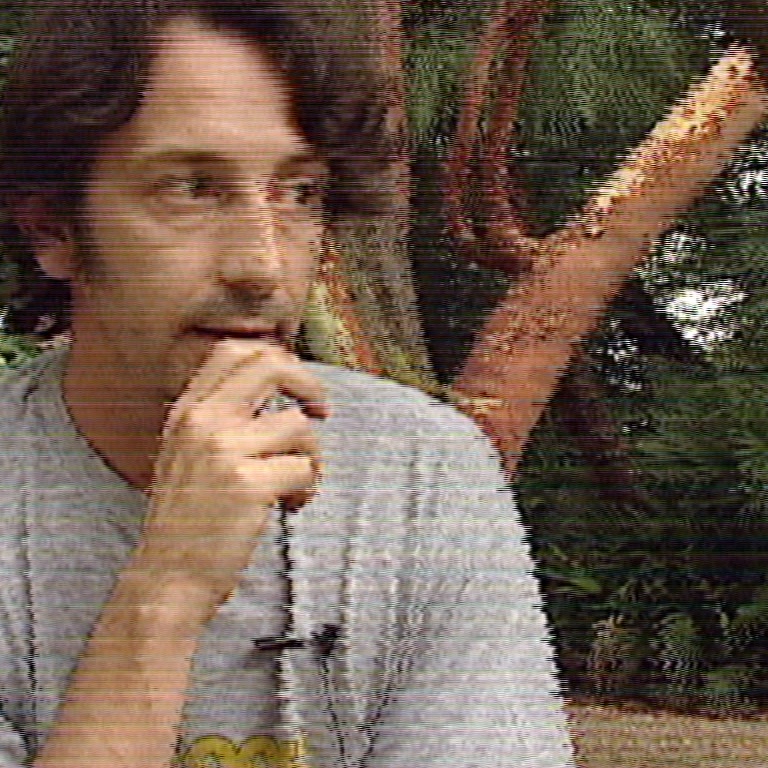}
    \caption{Frame 1}
    \label{fig:introduction_frame_b}
  \end{subfigure}
  \hfill
  \begin{subfigure}{0.32625\linewidth}
    \includegraphics[width=\linewidth]{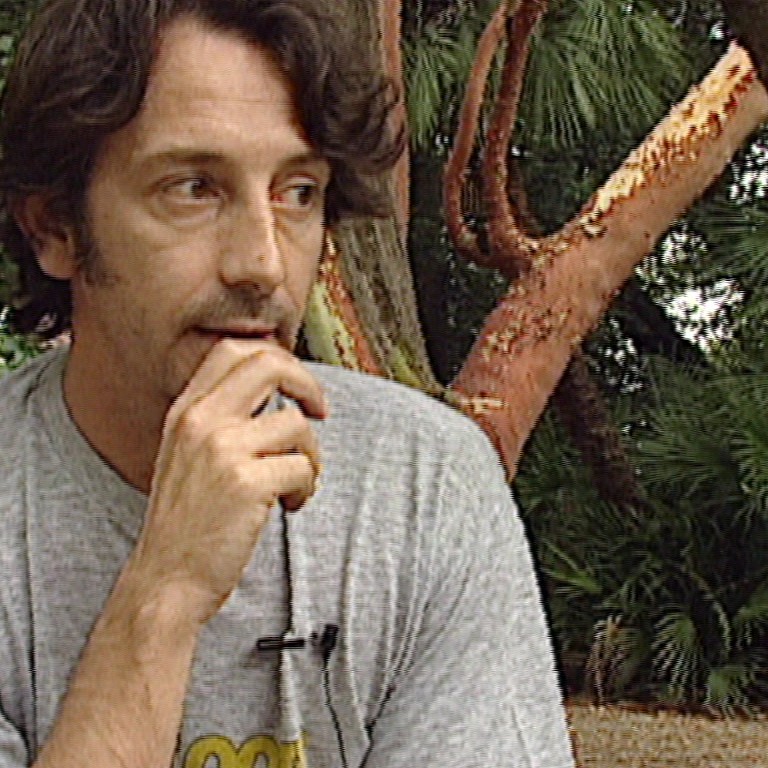}
    \caption{Frame 2}
    \label{fig:introduction_frame_c}
  \end{subfigure}
  \caption{Three consecutive frames of a real-world archival analog video. The degradations have significantly different intensities and positions and the temporal consistency is completely lost, which shows the time-varying nature of the artifacts. The third frame is essentially clean and could serve as a reference for the restoration.} \vspace{-1.5ex}
  \label{fig:introduction_frames}
\end{figure}

Analog magnetic tapes were widely used for video data storage from the 1950s to the late 1990s, and are still found in numerous archives worldwide with inestimable cultural value.
The aging of recording supports and the malfunctioning of reader devices introduce several types of artifacts that are unfortunately very common for analog videotapes, such as drop out, tape crease, scanline flickering, and tape mistracking \cite{stanco2013detection, van1996can}. These degradations limit the distribution and consumption of content by the general public if they are not properly restored. 
Nowadays, the restoration is usually performed frame-by-frame by experienced archivists using specialized commercial solutions, hence at great economic and time cost.
Since restoring a few hours of content can take up to weeks of work, applying this process to entire archives is essentially unfeasible. For this reason, the development of methods for the automatic restoration of analog videos is a necessity.

Standard video restoration works \cite{yu2022memory, ji2022multi, lin2022flow, liang2022vrt, liang2022recurrent} are designed for digital videos and do not consider the artifacts caused by media issues, which are more peculiar, more severe, and may occur simultaneously (see \cref{fig:introduction_frame_a}). Additionally, due to the severity of the degradation, the temporal consistency between the video frames is completely lost, which makes optical flow-based frame alignment techniques commonly used in video restoration \cite{zhou2022revisiting, chan2022basicvsr++} detrimental. Old video restoration methods consider the degradation related to the recording medium but focus on structured defects such as scratches \cite{iizuka2019deepremaster, wan2022bringing} or do not take advantage of the characteristics of the artifacts of analog videos \cite{stanco2016tracking, agnolucci2022restoration}. For instance, Agnolucci \etal \cite{agnolucci2022restoration} tackles analog video restoration with a transformer-based architecture that simply exploits the spatio-temporal information of input neighboring frames, thus without designing any strategy tailored for the task. In contrast, we propose an approach for analog video restoration that aims to remove the degradation caused by media issues -- especially the most severe ones, such as tape mistracking -- by leveraging their temporally varying nature.

\begin{figure*}
    \centering
    \includegraphics[width=\linewidth]{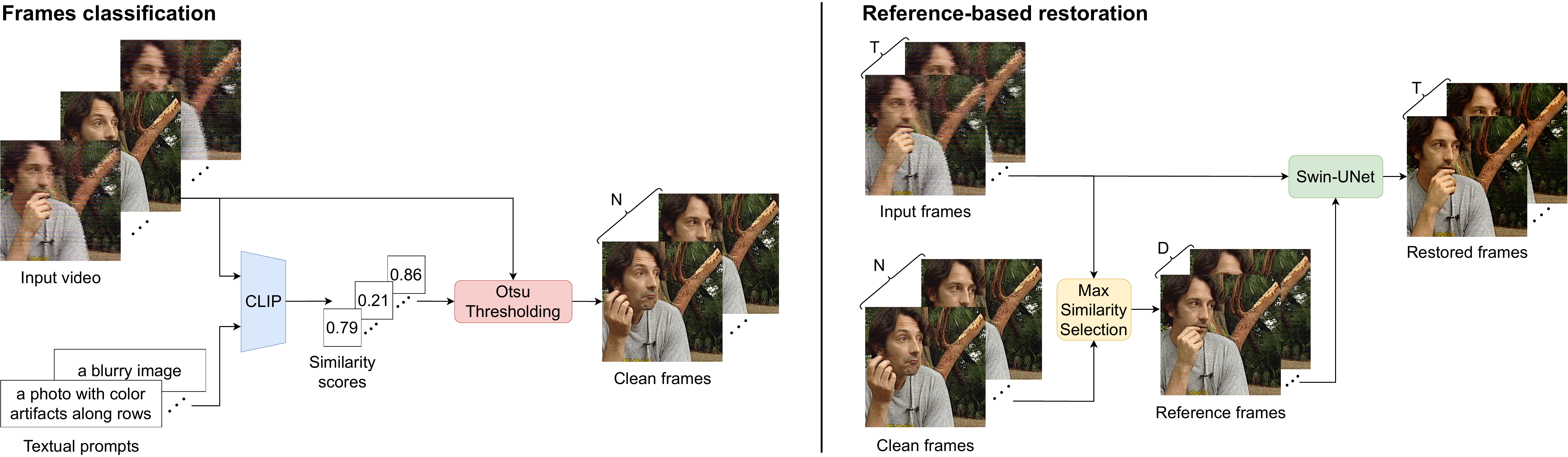}
    \caption{Overview of the proposed approach, named \method. \textit{Left:} given a video, we identify the cleanest frames with CLIP. First, we measure the similarity between the frames and textual prompts that describe different artifacts. Then, we employ Otsu's method to define a threshold for classifying the frames based on their similarity scores, resulting in a set of clean frames. \textit{Right:} given a window of $T$ degraded input frames, we select the most similar $D$ clean frames based on the CLIP image features and employ them as references. The proposed Swin-UNet then restores the input frames while effectively leveraging the references.}
    \label{fig:teaser}
\end{figure*}

\Cref{fig:introduction_frames} shows three consecutive frames that come from a real-world analog archive video. In the first two frames, the artifacts differ greatly in intensity and location, resulting in a complete loss of temporal consistency. In contrast, the last frame is essentially clean, so it contains a lot of high-quality details that could be useful for restoring temporally distant but similar degraded frames. This example shows the time-varying nature of the degradation in analog videos and motivates the main idea proposed in this work: identify the least damaged frames of a video and employ them as references for the restoration. To this end, we propose employing CLIP \cite{radford2021learning} for frame classification through zero-shot artifact detection by measuring the similarity between each frame and multiple textual prompts corresponding to different types of degradation. Then, within the frames classified as clean, we select those most similar to the input ones and employ them as references. We present a transformer-based Swin-UNet architecture that restores multiple input frames at once by exploiting the spatio-temporal information of neighboring frames while taking advantage of the references through our Multi-Reference Spatial Feature Fusion (MRSFF) block. MRSFF blocks employ multi-reference spatial cross-attention and attention pooling to restore the high-quality details that are lost in the input frames but still present in the cleaner reference ones. \Cref{fig:teaser} shows an overview of the proposed approach, named \method (\methodextended).
To overcome the lack of ground truth in real-world videos, we introduce a synthetic dataset of videos with artificially generated artifacts that emulate the degradation commonly found in analog ones. We hope that releasing our dataset will stimulate the research on analog video restoration to help preserve decades of video archives. Experimental results show the effectiveness of \method for both synthetic and real-world videos and its superior performance compared to state-of-the-art video restoration methods.

Our contributions can be summarized as follows:
\begin{itemize}
    \item We propose \method, a reference-based approach for analog video restoration that exploits the time-varying nature of the artifacts typical of this media by exploiting the cleanest frames of a video to restore the degraded ones;
    \item We propose to leverage CLIP for zero-shot artifact detection to identify the clean frames and select the references based on the similarity to the input ones;
    \item We develop a Swin-UNet architecture that takes advantage of the reference frames through our Multi-Reference Spatial Feature Fusion blocks;
    \item Quantitative and qualitative experiments on synthetic and real-world videos prove the effectiveness of our 
    method when compared with state-of-the-art video restoration techniques.
    
\end{itemize}

%% file: sec/2_related.tex
\section{Related Work}

\paragraph{Video Restoration}
Video restoration comprises various tasks that aim to enhance video quality, such as super-resolution and deblurring, each with its unique characteristics and methods \cite{lin2022flow, cao2021video, suin2021gated, maggioni2021efficient, geng2022rstt}. For example, MANA \cite{yu2022memory} tackles video super-resolution through cross-frame non-local attention and a memory bank learned during training, while \cite{ji2022multi} proposes a multi-scale memory-based architecture for video deblurring.
Other works propose approaches that achieve promising performance on multiple tasks \cite{chan2022basicvsr++, liang2022vrt, liang2022recurrent, zhou2022revisiting}. For example, RVRT \cite{liang2022recurrent} presents a recurrent transformer network with guided deformable attention, whereas BasicVSR++ \cite{chan2022basicvsr++} proposes a framework with second-order grid propagation and flow-guided deformable alignment. Despite the variety of degradations that are considered, these restoration techniques are designed for digital videos and do not address artifacts due to media issues typical of old ones.

\vspace{-0.8em}\paragraph{Old Video Restoration}
Restoration of old videos must address degradation and issues related to the recording medium.
Most of the existing methods focus on structured defects such as scratches and cracks typical of old films. Traditional approaches \cite{kim2006efficient, hongying2009efficient, saito1999image, giakoumis2005digital} rely on a detection network based on handcrafted features followed by an inpainting pipeline. More recently, deep learning-based methods have been proposed. DeOldify \cite{antic2018deoldify} is an open-source tool to restore old films based on a NoGAN training approach that is commonly used to colorize grayscale videos. \cite{iizuka2019deepremaster} presents a fully 3D convolutional for old video restoration and colorization. \cite{wan2022bringing} develops RTN, a recurrent transformer network with a hidden state that improves the temporal consistency of the results. Although impressive results are observed for structured defects, none of these techniques are able to correct such large and strong artifacts as tape mistracking.

To the best of our knowledge, only a few works aim to remove the typical degradations of analog videotapes. \cite{stanco2013detection, stanco2016tracking} exploit handcrafted features to detect artifacts and restore the corrupted tracking lines of interlaced videos by replacing them with those of adjacent frames. The most similar to our work is Agnolucci \etal \cite{agnolucci2022restoration}, which restores analog videos with a Swin-UNet network that leverages the spatio-temporal information of neighboring input frames. Therefore, \cite{agnolucci2022restoration} proposes a generic method without any specific strategy developed to exploit the intrinsic characteristics of the degradation. In contrast, our method is tailored for analog video restoration and takes advantage of the time-varying nature of the artifacts. Indeed, we first identify the cleanest frames of a video with CLIP and then leverage them as references through our MRSFF blocks.

%% file: sec/3_method.tex
\section{Proposed Approach}

\subsection{Synthetic Dataset} \label{sec:synthetic_dataset}
\textit{Archivio Storico Luce}, renowned as the largest Italian historical video archive and one of the largest in the world, houses an extensive collection of material spanning the entirety of the 1900s, sourced from various origins. The curators provided us with some degraded analog archival videos. In the following, we refer to these videos as ``real-world dataset", which consists of 4303 frames. Since these videos are obtained from the only available copy of the archive, they do not have a ground-truth counterpart and can not be used for training. Therefore, we create a synthetic dataset as similar as possible to real-world videos to train our model. Starting from high-quality videos belonging to the Harmonic dataset \cite{Harmonic-2019}, we employ Adobe After Effects \cite{christiansen2013adobe} to recreate different types of degradations \cite{hawkes2018identifying, AVAA}, which are visible in \cref{fig:introduction_frames}. In particular, we focus on: 1) tape mistracking and VHS edge waving, responsible for the horizontal displacement artifacts. These are the most complex distortions as they are the main cause of the temporal inconsistency; 2) chroma loss along the scanlines, visible from the horizontal cyan, magenta, and green lines; 3) tape noise, which is similar to Gaussian noise; 4) undersaturation, which makes the color of the frames look dull. We add these artifacts with random combinations, as well as positions and intensities, to each frame of the videos, purposely not to preserve temporal consistency. Indeed, as \cref{fig:introduction_frames} shows, in real-world videos the degradations occur at the same time and change abruptly between consecutive frames. Since we start from the same real-world videos, our pipeline for generating synthetic artifacts resembles that of \cite{agnolucci2022restoration}. However, contrary to \cite{agnolucci2022restoration}, we intend to release our synthetic dataset, in the hope of fostering further research on this task. In the end, we have 26,392 frames corresponding to 40 clips, which we divide into training and test sets with a 75\%-25\% ratio. More details on the synthetic dataset and a qualitative comparison with the real-world dataset are provided in \cref{sec:synthetic_dataset}.

\subsection{Frame Classification and Reference Selection} \label{sec:references_classification}

\begin{figure*}
    \centering
    \includegraphics[width=0.9\linewidth]{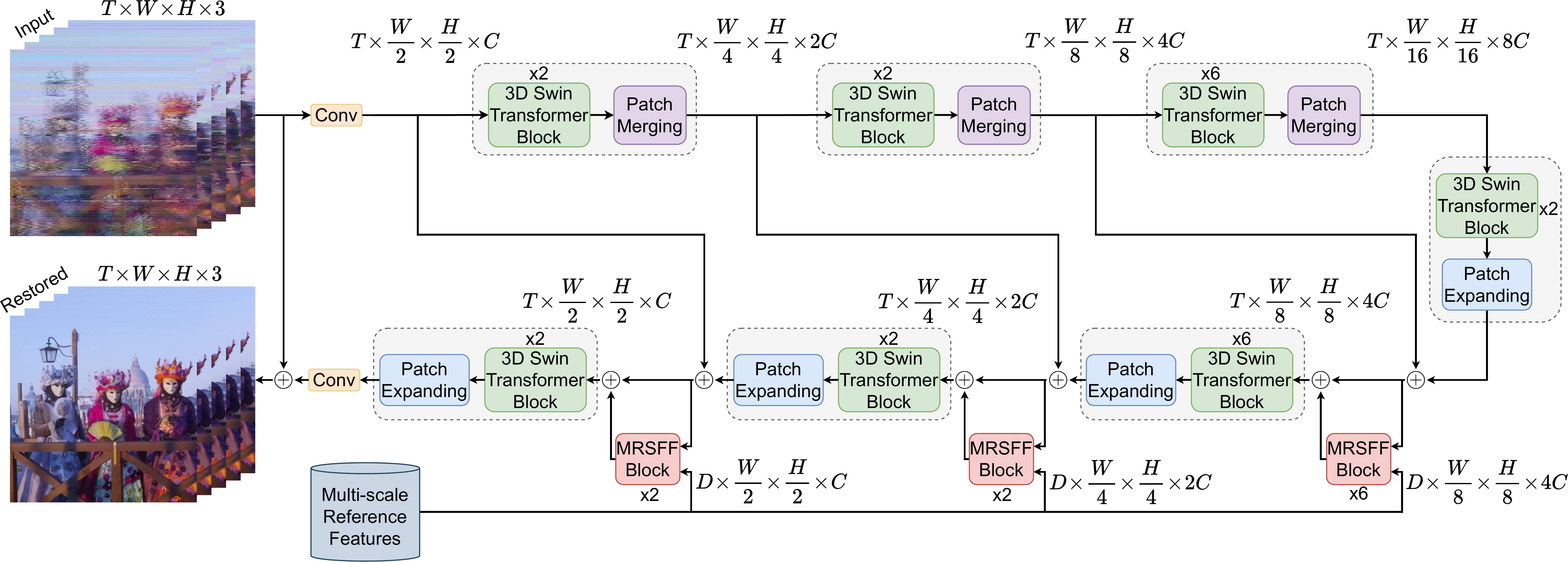}
    \caption{Overview of the proposed Swin-UNet architecture. We extract multi-scale features from the reference frames through a pre-trained Swin Transformer feature extractor. Then we exploit them through the proposed MRSFF blocks in the decoding part of the UNet.}
    \label{fig:architecture}
\end{figure*}

Given the time-varying nature of the artifacts in analog videos, we can identify the cleanest (\ie least degraded) frames of a video to exploit them as references for restoring the other frames. In other words, we can classify the frames into fairly clean (\ie the possible references) and very degraded ones. A possible approach could be training a binary classifier, but that would introduce two disadvantages: 1) we would have to choose a predefined threshold over which we consider a frame clean, which is an ill-posed problem; 2) all frames of a video could be classified as degraded, meaning that we would have no references. Hence, a continuous no-reference metric is more suitable for this task because it would allow one to define a different threshold for each video in an unsupervised manner.

We rely on CLIP \cite{radford2021learning} for our purpose. CLIP is a multimodal model trained with an image-caption alignment objective that obtained remarkable results in several downstream tasks, such as image generation \cite{galatolo2021generating}, retrieval \cite{baldrati2022conditioned,baldrati2023zero}, and quality assessment \cite{wang2022exploring, zhang2023blind}. Given a video, we propose to employ CLIP for zero-shot artifact detection by measuring the similarity between each frame and a textual prompt describing some kind of artifact (\eg ``\textit{an image with color artifacts along rows}"). Therefore, a lower similarity corresponds to a less degraded frame. To improve the robustness of the predictions, we combine prompts referring to different types of artifacts with prompt ensembling \cite{radford2021learning} by averaging CLIP text features. We choose to use prompts that identify degraded frames because the visual artifacts they present are more easily described in natural language compared to the unconstrained domain of the content of clean ones. To adapt our method to address new types of degradation, it is enough to update the list of prompts, \eg to specialize the approach for specific videos or types of medium. We provide more details on the list of prompts in \cref{sec:list_prompts}. Then, we build a histogram of the CLIP similarities of all the frames. Intuitively, we expect this histogram to be bimodal, with the two peaks corresponding to degraded and almost clean frames. To define a threshold to classify the frames into two classes, we propose using Otsu's method \cite{otsu1979threshold}, which is commonly adopted for automatic image thresholding \cite{sezgin2004survey}. The algorithm returns a single threshold determined by minimizing the intra-class variance in an unsupervised manner, thus based on the histogram of the similarities of each individual video. We classify as clean each frame with a CLIP similarity score lower than the threshold. In this way, we obtain a set of $N$ clean frames from the video, which we can leverage as references for the restoration. The number of clean frames $N$ changes for every video, as it is not a hyperparameter but rather depends on the degradation to which the video is subjected. The left-hand side of \cref{fig:teaser} shows an overview of the frame classification process.

Finally, given a window of $T$ input neighboring frames, we compute the cosine similarity of the CLIP image features between the central frame and the set of clean ones. Then, we take the $D$ most similar clean frames and use them as references. In our experience, artifacts in real-world videos never severely degrade all frames, so we always have significantly more than $D$ clean frames to choose from (see the supplementary material for more details).
In other words, $N$ always exceeds $D$.
Note that the arrangement of the reference frames is arbitrary as they are not necessarily temporally neighboring. The right section of \cref{fig:teaser} shows the reference frames selection process and the subsequent reference-based restoration.

\subsection{Swin-UNet Architecture}

\Cref{fig:architecture} shows the proposed reference-based video restoration network. As in \cite{agnolucci2022restoration}, our architecture is based on a Swin-UNet that restores a window of $\mathit{T}$ degraded frames at once. Since the complexity of attention is quadratic to the number of elements within the attention window, global attention on the entire input frames is unfeasible. Hence, we leverage the Swin Transformer \cite{liu2021swin, liu2022video} to reduce the computational cost while maintaining the ability to learn long-range dependencies. Indeed, the Swin Transformer computes global attention only within local windows and then enables cross-window connections with a shifted-window mechanism. In this way, we can rely on transformers and their attention mechanism efficiently, in order to also take advantage of their ability to deal with frames that are not perfectly aligned. Indeed, due to the strong horizontal displacements caused by the artifacts, we cannot correctly align the input frames with the combination of explicit motion estimation and image warping. However, it has been shown that the expressiveness of the transformers alleviates this problem \cite{liang2022vrt, wan2022bringing}.

First, we extract shallow features from the input with a convolutional layer. Then, in the encoder, we reduce the patch size and increase the number of channels through cascaded Swin 3D transformer blocks and patch merging layers \cite{liu2022video}. In the decoder, we elaborate the processed features (\ie those that go through the encoder and decoder) via Swin 3D transformer blocks and patch expanding layers, which consist of pixel shuffle layers. We use skip connections to add the encoder residual features to the processed ones. A skip connection between the input and the output makes the network learn the residual of each frame.

Using Swin 3D transformer blocks allows taking advantage of the spatio-temporal information of the input frames thanks to self-attention. Indeed, by having the processed features attend themselves, we intuitively make each region of the input frames look at similar parts of the other frames. This is particularly useful due to the time-varying nature of the artifacts, as a highly degraded portion of one frame may be less damaged in one of the neighboring ones. However, if a given region is severely degraded in all the input frames, some details will be permanently lost. For this reason, we propose employing clean reference frames that do not belong to the window of the input frames. Contrary to \cite{agnolucci2022restoration}, our model also takes as input $\mathit{D}$ reference frames -- selected as explained in \cref{sec:references_classification} -- from which we extract multi-scale features with a pre-trained frozen Swin Transformer feature extractor. In the decoder, the proposed MRSFF blocks combine processed and reference features to recover details that would be lost otherwise.

\subsection{Multi-Reference Spatial Feature Fusion Block} \label{sec:MRSFF_block}

\begin{figure}[]
    \centering
    \includegraphics[width=0.7\columnwidth]{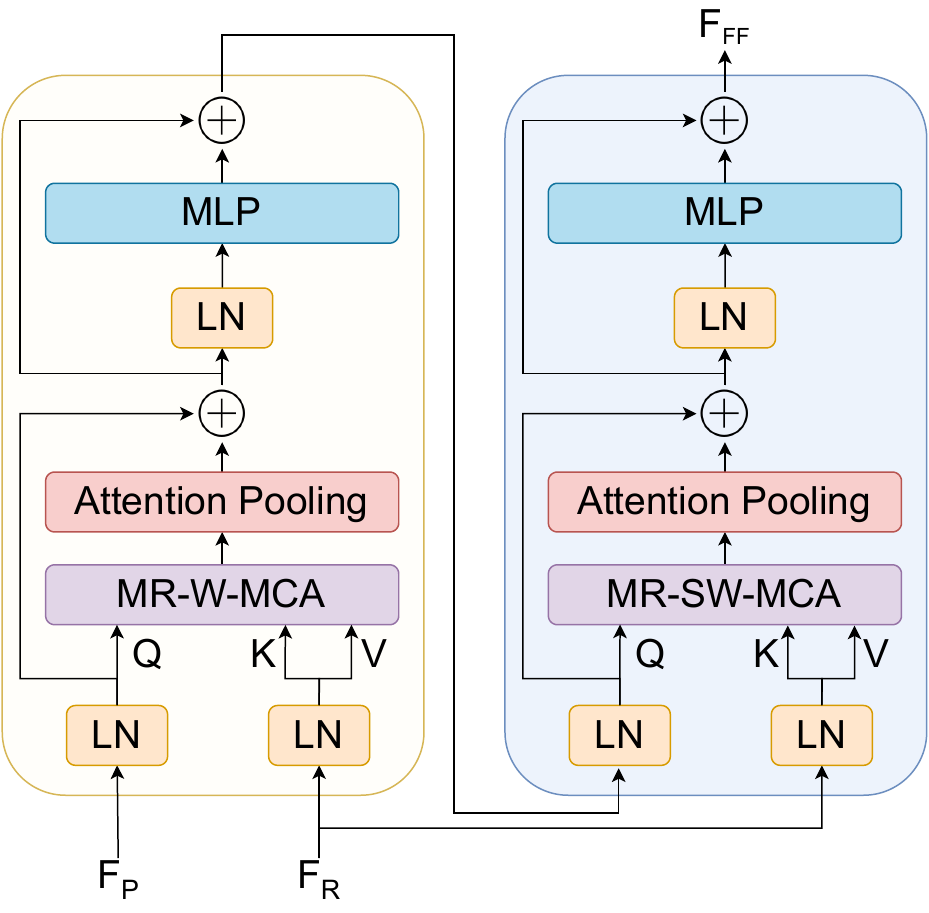}
    \caption{Two successive MRSFF blocks. LN stands for Layer Normalization. MR-(S)W-MCA represents the Multi-Reference-(Shifted)Window-Multi-head Cross Attention.}
    \label{fig:block}
\end{figure}

Given the reference features, our aim is to exploit them for the restoration. Since the reference frames are not contiguous in time (see \cref{sec:references_classification}), the use of temporal attention with Swin 3D transformer blocks is not suitable because it assumes some sort of temporal correlation. Furthermore, due to the variability in artifact locations across input frames, we want the processed features to independently attend to the reference ones along the spatial dimension. In this way we allow the processed features to focus on recovering the missing details in the corresponding input frames. For these reasons, we present our Multi-Reference Spatial Feature Fusion block, which makes use of Multi-Reference-(Shifted)Window-Multi-Head Cross Attention (MR-(S)W-MCA) and attention pooling. MRSFF blocks are inspired by Swin 2D transformer blocks \cite{liu2021swin}. Indeed, both architectures rely on a shifted window-based attention mechanism along the spatial dimension. However, while Swin 2D uses self-attention, MRSFF blocks perform cross-attention between two inputs that may have different dimensions, since the number of input and reference frames T and D may not coincide. In addition, MRSFF blocks employ attention pooling to combine the information coming from the reference frames.

Similarly to \cite{liu2021swin}, we partition each feature $\mathit{F} \!\in\! \mathbb{R}^{H \times W \times C}$ into $\frac{\textit{H}\textit{W}}{\textit{M}^2}$ non-overlapping $\mathit{M} \!\times\! \mathit{M}$ windows. Here, $\mathit{H}$ and $\mathit{W}$ are the height and width, respectively, and $\mathit{C}$ is the number of channels. We flatten the $\mathit{M}^2$ elements of each window and compute the local attention.  In the successive MRSFF block, a $\left \lfloor \frac{\textit{M}}{2} \right \rfloor \times \left \lfloor \frac{\textit{M}}{2} \right \rfloor$ cyclic spatial shift \cite{liu2021swin} of the features before window partitioning enables cross-window connections. MR-(S)W-MCA is based on a cross-attention mechanism, in which the processed features associated with each input frame act as independent queries, while the reference features serve as both keys and values. Intuitively, each input frame looks at similar parts of the reference frames and then exploits them to recover its missing details. \Cref{fig:block} shows two consecutive MRSFF blocks. 

Formally, let $\mathit{F}_P \in \mathbb{R}^{T \times M^2 \times C}$ and $F_{R} \in \mathbb{R}^{D \times M^2 \times C}$ be, respectively, the processed and reference features of a given local window. For each head, we compute:
\begin{align}
    &\mathit{Q}_P = \mathit{F}_P \mathit{P}_Q \in \mathbb{R}^{T \times M^2 \times C} \nonumber \\
    &\mathit{K}_R = \mathit{F}_R \mathit{P}_K, \mathit{V}_R = \mathit{F}_R \mathit{P}_V \in \mathbb{R}^{D \times M^2 \times C}
    \label{eq:qkv}
\end{align}
where $\mathit{P}_Q$, $\mathit{P}_K$, $\mathit{P}_V$ $\in \mathbb{R}^{C \times C}$ are projection matrices. Let $\mathit{Q}_{P_i}$, $\mathit{K}_{R_j}$, $\mathit{V}_{R_j}$ $\in \mathbb{R}^{M^2 \times C}$ be the \textit{i}-th and  \textit{j}-th elements of the queries, keys and values, respectively.
We compute the attention map $\mathit{A_{ij}} = SoftMax(\mathit{Q}_{P_i} \mathit{K}^{T}_{R_j} / \sqrt{C} + \mathit{B}) \in \mathbb{R}^{M^2 \times M^2}$. As in \cite{liu2021swin}, we include a learnable relative positional embedding $\mathit{B} \!\in\! \mathbb{R}^{M^2 \times M^2}$ in each attention head.
Conceptually, $\mathit{A_{ij}}$ represents the correlation between the \textit{i}-th input frame and the \textit{j}-th reference frame and is used for a weighted sum of $\mathit{V}_{R_j}$, formulated as $\mathit{A_{ij}} \mathit{V}_{R_j} \in \mathbb{R}^{M^2 \times C}$. For each input frame \textit{i} $\in \{1, \ldots, \mathit{T} \}$, we repeat this computation for all the $\mathit{D}$ references  and concatenate the results:
\begin{equation}
    \mathit{F}_{G} = \text{MR-(S)W-MCA}(\mathit{F}_P, \mathit{F}_R) \in {\mathbb{R}^{T \times D \times M^2 \times C}}
    \label{eq:final_attention_formula}
\end{equation}
Intuitively, the features pertaining to each of the $T$ frames represent the fusion of the information from the corresponding input frame with that of the $D$ reference ones. Consequently, it is imperative to effectively combine these features on the $D$ axis to fully leverage the reference frames. Drawing inspiration from CLIP \cite{radford2021learning}, we propose to use attention pooling to achieve this goal. Attention pooling relies on the self-attention mechanism to select the most pertinent features for a pooled representation, rather than just taking the average. In other words, attention pooling allows us to take advantage of the most valuable parts of the information coming from the reference frames to obtain an aggregate representation. 
Note that CLIP uses attention pooling to reduce the 3D feature maps of a CNN into single vectors, thus removing the height and width dimensions. Our aim, on the contrary, is to aggregate $D$ sequences of transformer tokens into a single one, thereby pooling along a single axis. As in \cite{radford2021learning}, we condition the query on the global average-pooled representation. First, we compute the average of the features along the $D$ axis:
\begin{equation}
    \mathit{F}_{avg} = \text{Avg}(\mathit{F}_{G}) \in {\mathbb{R}^{T \times M^2 \times C}}
\end{equation}
Then, we concatenate the $D$ sequences of tokens:
\begin{equation}
    \mathit{F}_{C} = \text{Concatenation}(\mathit{F}_{G}) \in {\mathbb{R}^{T \times DM^2 \times C}}
\end{equation}
Finally, for each input frame, we project the features and rely on the attention mechanism by considering the average features $\mathit{F}_{avg}$ as the query and the concatenated ones $\mathit{F}_{C}$ as both the keys and values:
\begin{equation}
    \mathit{F}_{FF} = \text{MHA}(\mathit{F}_{avg}, \mathit{F}_{C}, \mathit{F}_{C}) \in {\mathbb{R}^{T \times M^2 \times C}}
\end{equation}
where MHA represents both the projection and multi-head attention operations. In this way, we obtain an aggregate representation of the features that encapsulate the information from the reference frames. Then, we further transform the features via a layer normalization layer and an MLP. In the successive MRSFF block, the fused features $\mathit{F}_{FF}$ serve as the queries of the transformer, assuming the role held by the processed features $\mathit{F}_P$ in the preceding block. Nevertheless, in this case, the features are shifted before window partitioning to enable cross-window connections and therefore learn long-range dependencies.

Ultimately, the fused $\mathit{F}_{FF}$ and the processed $\mathit{F}_{P}$ features of each window share the same dimensions and can consequently be added together to recover the lost details.

%% file: sec/4_results.tex
\section{Experimental Results}

\subsection{Implementation Details}
We train the proposed model on the synthetic dataset for 100 epochs using the ADAMW optimizer \cite{loshchilov2017decoupled} with weight decay equal to 0.01 and $(\beta_1, \beta_2) = (0.9, 0.99)$. The learning rate is set to $2e-5$. We use a frozen CLIP RN50x4 model for frame classification. During training, we randomly crop $128 \times 128$ patches from the frames. We set the number of input frames $\mathit{T}$ to 5; the number of reference frames $\mathit{D}$ to 5; $\mathit{M}$ and $\mathit{C}$ in \cref{sec:MRSFF_block} to 8 and 96. We train \method with a weighted sum of the Charbonnier loss \cite{charbonnier1994two} and a perceptual loss \cite{johnson2016perceptual, ledig2017photo, dosovitskiy2016generating}. More details on the training loss are reported in \cref{sec:training_loss}. During testing, we crop the $512 \times 512$ center patch of each frame. Quantitative results are the average over the videos of the test set. On a single A100-40GB GPU, our model takes 2 days for training and runs at 15FPS at inference time, which is more than suitable for a task with no real-time requirements. 

\subsection{Baselines}
We compare \method with the following baselines: 1) MANA \cite{yu2022memory}: a memory-augmented architecture with cross-frame non-local attention for video super-resolution; 2) MemDeblur \cite{ji2022multi}: a multi-scale memory-augmented recurrent architecture for video deblurring; 3) BasicVSR++ \cite{chan2022basicvsr++}: a recurrent framework with second-order grid propagation and flow-guided deformable alignment for video restoration; 4) RVRT \cite{liang2022recurrent}: a recurrent transformer with guided deformable attention for video restoration; 5) RTN \cite{wan2022bringing}: a recurrent transformer network for old film restoration; 6) Agnolucci \etal \cite{agnolucci2022restoration}: a Swin-UNet architecture for analog video restoration.

The baselines comprise standard and old video restoration works to show the difference from analog video restoration. For a fair comparison, we trained all the baselines from scratch on our training dataset using the official repositories.

\subsection{Evaluation Metrics}
We evaluate the performance on the synthetic dataset using four full-reference metrics: two signal-based metrics, PSNR and SSIM \cite{wang2004image}, to measure the low-level difference between restored and ground-truth frames; LPIPS \cite{zhang2018unreasonable}, which better correlates with human perceived visual quality; VMAF \cite{li2016toward}, a perceptual video quality assessment model that combines multiple elementary quality metrics, including one that accounts for the temporal difference between adjacent frames, thus evaluating the presence of motion jitter and flicker.

Since the ground truth is not available for the real-world dataset, we employ three no-reference image quality assessment metrics: BRISQUE \cite{mittal2012no}, NIQE \cite{mittal2012making} and CONTRIQUE \cite{madhusudana2022image}.

\subsection{Quantitative Results} \label{sec:quantitative_results}
We report the quantitative results for the synthetic and real-world datasets in \cref{tab:synthetic_quantitative_results,tab:real_world_quantitative_results}, respectively. Considering the synthetic dataset, \method outperforms all baselines on all metrics by a large margin. In particular, LPIPS shows that our method produces results that are more perceptually accurate than the other techniques, while VMAF proves that our restored videos are more temporally consistent and with less motion jitter. Furthermore, we observe that BasicVSR++ \cite{chan2022basicvsr++} and RVRT \cite{liang2022recurrent} perform poorly, even though they represent the state of the art in standard video restoration. We attribute this result to the use of optical flow for frame alignment, which is detrimental for analog videos, as the degradation is so severe that it completely disrupts the temporal consistency. This outcome further shows the difference between standard and analog video restoration.

Considering the real-world dataset, our approach achieves the best results for CONTRIQUE, while MANA \cite{yu2022memory} performs better for BRISQUE and NIQE. However, the qualitative results presented in \cref{sec:qualitative_results} show that MANA adds high-frequency artifacts to restored frames. We argue that BRISQUE and NIQE are misled by these artifacts and mistake them for high-frequency details that are instead typical of high-quality images \cite{seidenari2022language,agnolucci2023perceptual}. See \cref{sec:analysis_quantitative_results} for more details.

\begin{table}[]
  \centering
  \Huge
  \resizebox{\columnwidth}{!}{ 
  \begin{tabular}{lcccc}
    \toprule
    Method & PSNR$\:\uparrow$ & SSIM$\:\uparrow$ & LPIPS$\:\downarrow$ & VMAF$\:\uparrow$\\
    \midrule
    MANA \cite{yu2022memory} & 27.81 & 0.843 & 0.206 & 40.28 \\
    MemDeblur \cite{ji2022multi} & 33.22 & 0.911 & 0.106 & 71.55 \\
    BasicVSR++ \cite{chan2022basicvsr++} & 31.66 & 0.916 & 0.098 & 78.91 \\
    RVRT \cite{liang2022recurrent} & 32.47 & 0.896 & 0.117 & 72.41 \\
    RTN \cite{wan2022bringing} & 31.46 & 0.905 & 0.100 & 56.76 \\
    Agnolucci \etal \cite{agnolucci2022restoration} & 34.96 & 0.940 & 0.060 & 77.83 \\ \midrule
    \rowcolor{tabhighlight}
    \textbf{\method} & \textbf{35.53} & \textbf{0.946} & \textbf{0.052} & \textbf{83.61} \\
    \bottomrule
  \end{tabular}
  }
  \caption{Quantitative results for the synthetic dataset. $\uparrow$ means that higher values are better, $\downarrow$ means that lower values are better. Best results are highlighted in bold.}
  \label{tab:synthetic_quantitative_results}
\end{table}

\begin{table}[]
  \centering
  \Huge
  \resizebox{\columnwidth}{!}{ 
  \begin{tabular}{lccc}
    \toprule
    Method & BRISQUE$\:\downarrow$ & NIQE$\:\downarrow$ & CONTRIQUE$\:\downarrow$\\
    \midrule
    MANA \cite{yu2022memory} & \textbf{41.80} & \textbf{5.90} & 48.18 \\
    MemDeblur \cite{ji2022multi} & 51.20 & 8.89 & 45.82 \\
    BasicVSR++ \cite{chan2022basicvsr++} & 59.19 & 8.42 & 48.44 \\
    RVRT \cite{liang2022recurrent} & 47.61 & 8.39 & 48.64 \\
    RTN \cite{wan2022bringing} & 53.27 & 6.94 & 46.17 \\
    Agnolucci \etal \cite{agnolucci2022restoration} & 59.44 & 7.90 & 45.45 \\ \midrule
    \rowcolor{tabhighlight}
    \textbf{\method} & 56.04 & 7.74 & \textbf{42.99} \\
    \bottomrule
  \end{tabular}
  }
  \caption{Quantitative results for the real-world dataset. $\downarrow$ means that lower values are better. Best results are highlighted in bold.}
  \label{tab:real_world_quantitative_results}
\end{table}

\begin{figure*}[!htb]
    \captionsetup[subfigure]{labelformat=empty}
    \centering
    
    \includegraphics[width=\syntheticwidth]{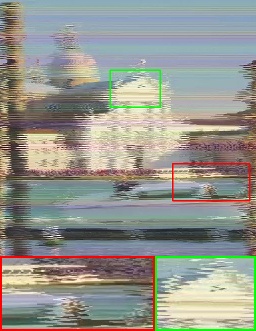}
    \hfill
    \includegraphics[width=\syntheticwidth]
    {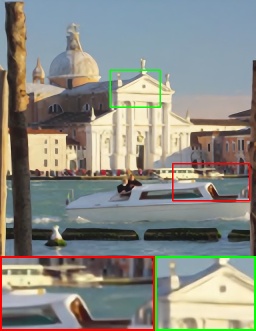}
    \hfill
    \includegraphics[width=\syntheticwidth]{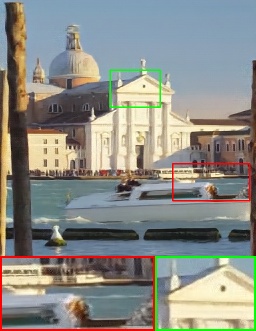}
    \hfill
    \includegraphics[width=\syntheticwidth]
    {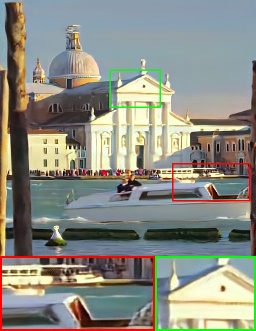}
    \hfill
    \includegraphics[width=\syntheticwidth]
    {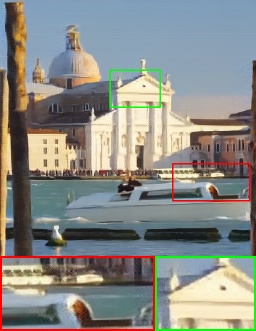}
    \hfill
    \includegraphics[width=\syntheticwidth]
    {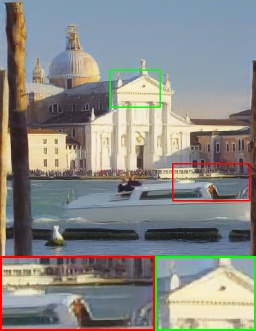}
    \hfill
    \includegraphics[width=\syntheticwidth]{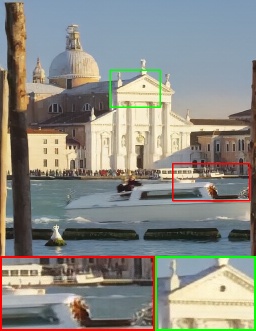}
    \hfill
    \includegraphics[width=\syntheticwidth]{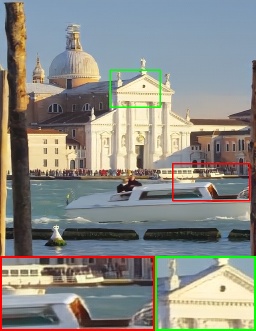}
    \hfill
    \includegraphics[width=\syntheticwidth]{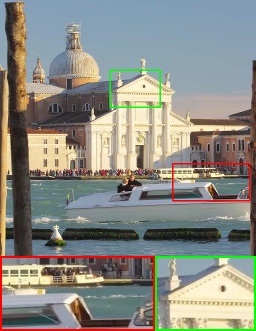}
    
    \includegraphics[width=\syntheticwidth]{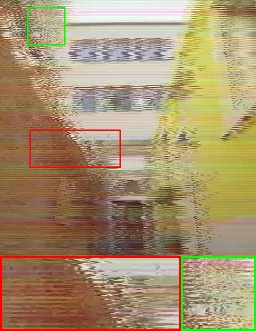}
    \hfill
    \includegraphics[width=\syntheticwidth]{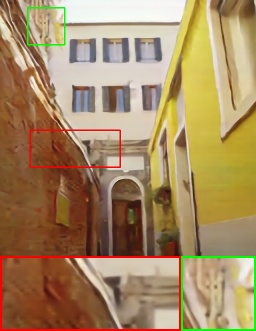}
    \hfill
    \includegraphics[width=\syntheticwidth]{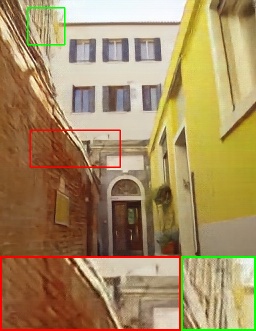}
    \hfill
    \includegraphics[width=\syntheticwidth]
    {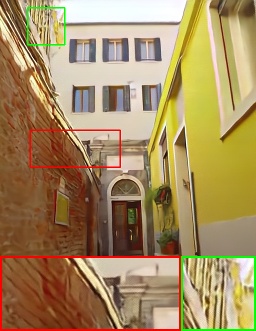}
    \hfill
    \includegraphics[width=\syntheticwidth]
    {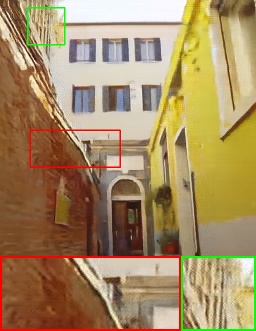}
    \hfill
    \includegraphics[width=\syntheticwidth]{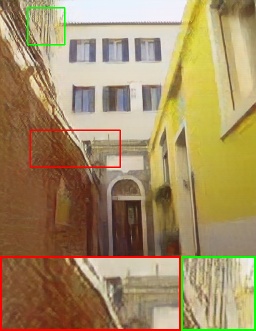}
    \hfill
    \includegraphics[width=\syntheticwidth]{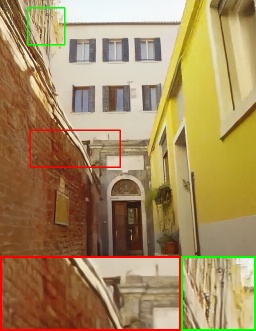}
    \hfill
    \includegraphics[width=\syntheticwidth]{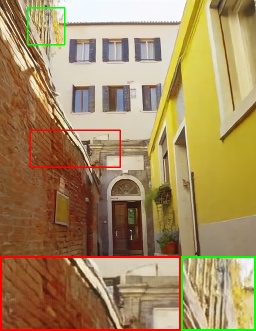}
    \hfill
    \includegraphics[width=\syntheticwidth]{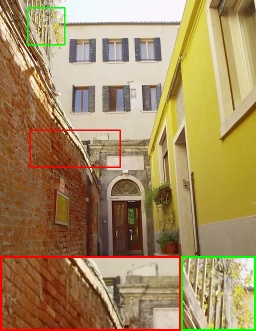}
    
    \begin{subfigure}{\syntheticwidth}
        \includegraphics[width=\linewidth]{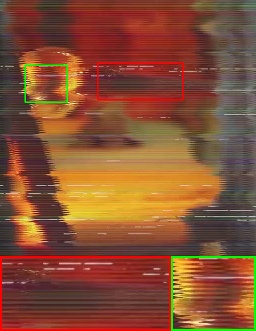}
        \caption{Input}
    \end{subfigure}
    \hfill
    \begin{subfigure}{\syntheticwidth}
        \includegraphics[width=\linewidth]{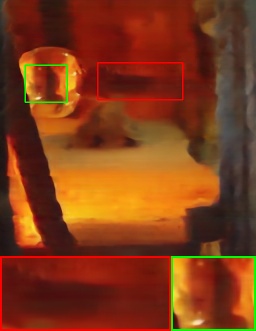}
        \caption{MANA}
    \end{subfigure}
    \hfill
    \begin{subfigure}{\syntheticwidth}
        \includegraphics[width=\linewidth]{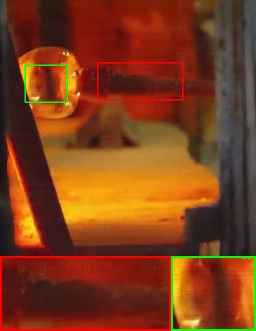}
        \caption{MemDeblur}
    \end{subfigure}
    \hfill
    \begin{subfigure}{\syntheticwidth}
        \includegraphics[width=\linewidth]{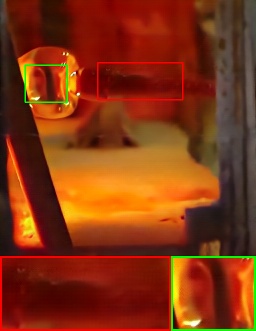}
        \caption{BasicVSR++}
    \end{subfigure}
    \hfill
    \begin{subfigure}{\syntheticwidth}
        \includegraphics[width=\linewidth]{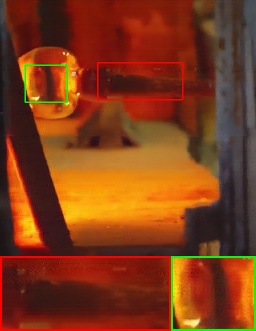}
        \caption{RVRT}
    \end{subfigure}
    \hfill
    \begin{subfigure}{\syntheticwidth}
        \includegraphics[width=\linewidth]{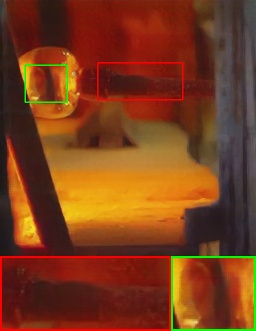}
        \caption{RTN}
    \end{subfigure}
    \hfill
    \begin{subfigure}{\syntheticwidth}
        \includegraphics[width=\linewidth]{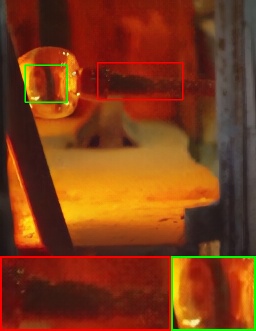}
        \caption{Agnolucci \etal}
    \end{subfigure}
    \hfill
    \begin{subfigure}{\syntheticwidth}
        \includegraphics[width=\linewidth]{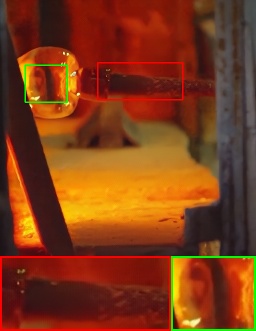}
        \caption{\textbf{\method}}
    \end{subfigure}
    \hfill
    \begin{subfigure}{\syntheticwidth}
        \includegraphics[width=\linewidth]{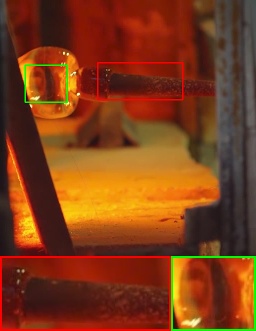}
        \caption{GT}
    \end{subfigure}
    \caption{Qualitative results for the synthetic dataset. Best viewed in PDF.}
    \label{fig:synthetic_qualitative_results}
\end{figure*}

\begin{figure*}[!htb]
    \captionsetup[subfigure]{labelformat=empty}
    \centering
    
    \includegraphics[width=\realworldwidth]{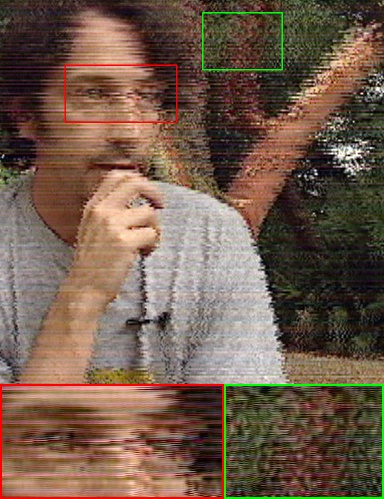}
    \hfill
    \includegraphics[width=\realworldwidth]{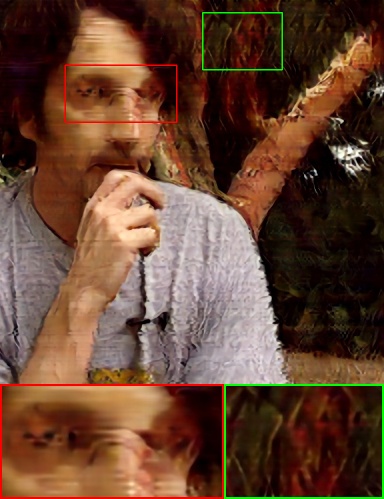}
    \hfill
    \includegraphics[width=\realworldwidth]{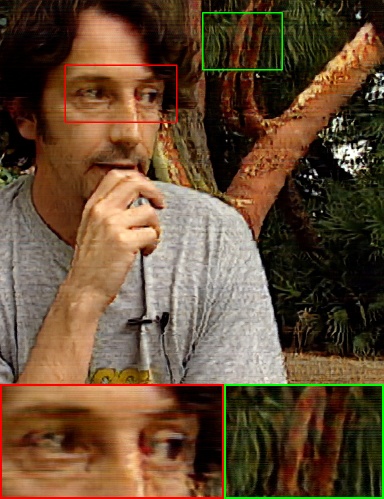}
    \hfill
    \includegraphics[width=\realworldwidth]{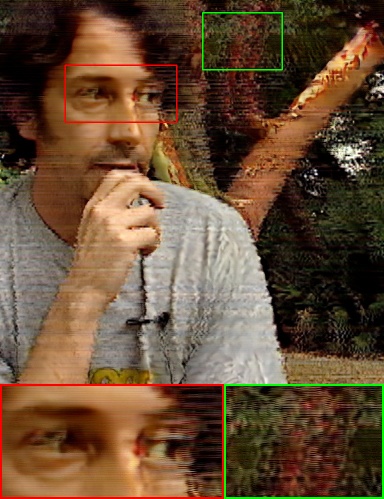}
    \hfill
    \includegraphics[width=\realworldwidth]{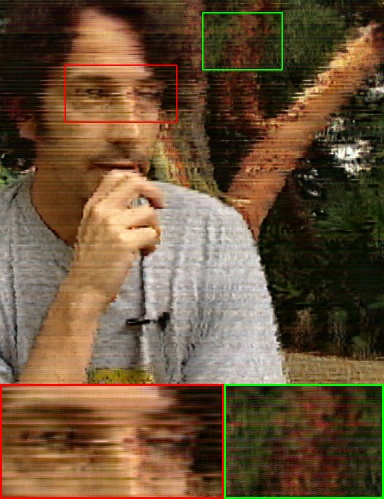}
    \hfill
    \includegraphics[width=\realworldwidth]{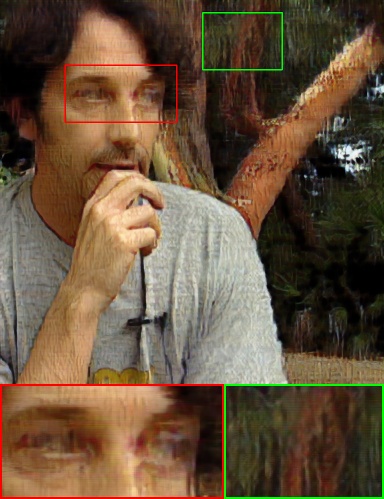}
    \hfill
    \includegraphics[width=\realworldwidth]{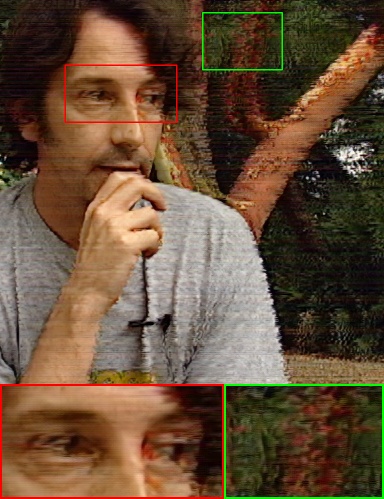}
    \hfill
    \includegraphics[width=\realworldwidth]{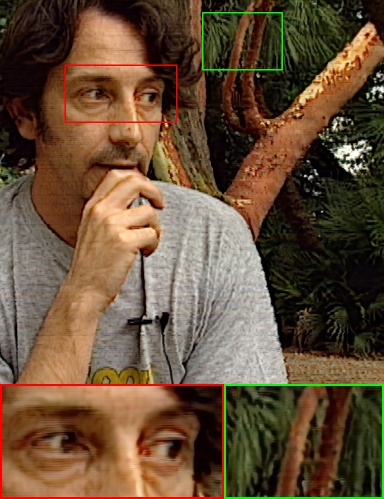}

    \begin{subfigure}{\realworldwidth}
        \includegraphics[width=\linewidth]{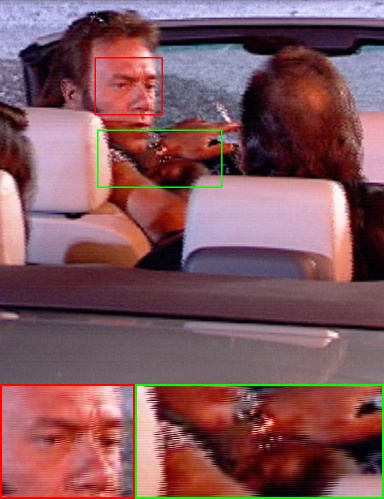}
        \caption{Input}
    \end{subfigure}
    \hfill
    \begin{subfigure}{\realworldwidth}
        \includegraphics[width=\linewidth]{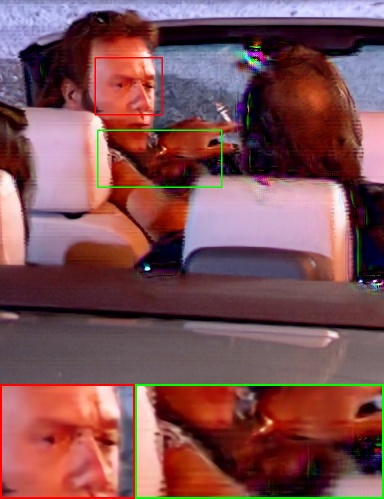}
        \caption{MANA}
    \end{subfigure}
    \hfill
    \begin{subfigure}{\realworldwidth}
        \includegraphics[width=\linewidth]{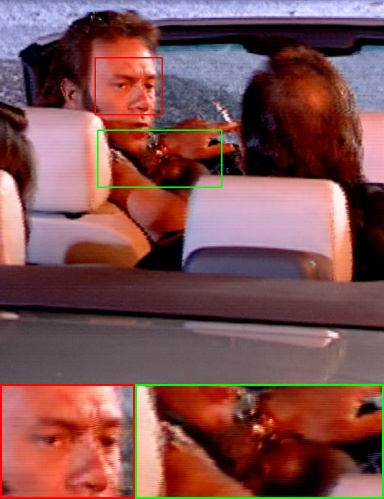}
        \caption{MemDeblur}
    \end{subfigure}
    \hfill
    \begin{subfigure}{\realworldwidth}
        \includegraphics[width=\linewidth]{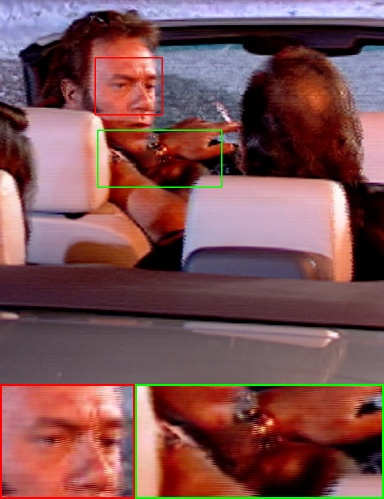}
        \caption{BasicVSR++}
    \end{subfigure}
    \hfill
    \begin{subfigure}{\realworldwidth}
        \includegraphics[width=\linewidth]{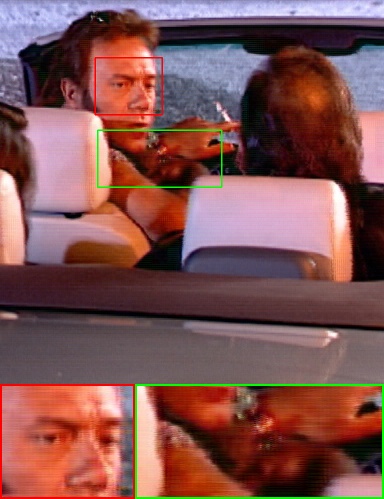}
        \caption{RVRT}
    \end{subfigure}
    \hfill
    \begin{subfigure}{\realworldwidth}
        \includegraphics[width=\linewidth]{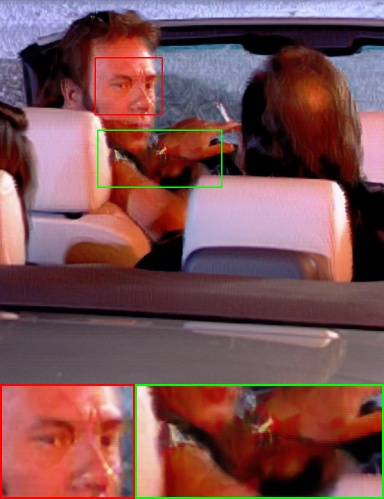}
        \caption{RTN}
    \end{subfigure}
    \hfill
    \begin{subfigure}{\realworldwidth}
        \includegraphics[width=\linewidth]{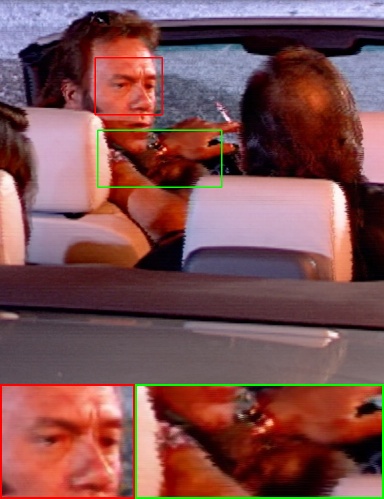}
        \caption{Agnolucci \etal}
    \end{subfigure}
    \hfill
    \begin{subfigure}{\realworldwidth}
        \includegraphics[width=\linewidth]{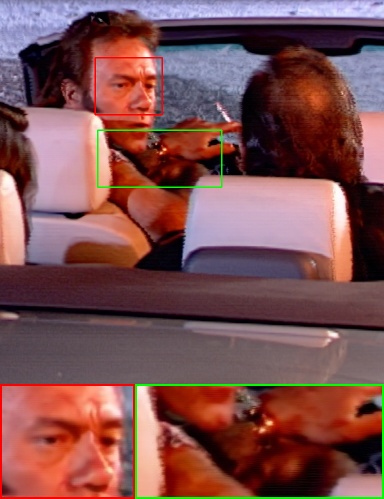}
        \caption{\textbf{\method}}
    \end{subfigure}
    \caption{Qualitative results for the real-world dataset. Best viewed in PDF.}
    \label{fig:real_qualitative_results}
\end{figure*}

\subsection{Qualitative Results} \label{sec:qualitative_results}
\Cref{fig:synthetic_qualitative_results} shows the qualitative results for the synthetic dataset. We observe that \method generates the most detailed and photorealistic images. MANA and BasicVSR++ yield results that lack photorealism and fine-grained details, while MemDeblur, RVRT, and RTN generate images that exhibit a higher degree of accuracy, albeit accompanied by a discernible amount of noise.
The results of Agnolucci \etal are satisfactory, but with fewer details compared to our approach. See for example the white boat in the first row or the brick wall in the second row.

We report qualitative results of the real-world dataset in \cref{fig:real_qualitative_results}. As mentioned in \cref{sec:quantitative_results}, despite the promising quantitative results, MANA and RTN generate clearly unsatisfactory images with many unpleasant artifacts, showing the unreliability of BRISQUE and NIQE for this task. RVRT proves to be unable to generalize to real-world videos, since it fails to remove most of the degradation from the input frames. MemDeblur, BasicVSR++, and Agnolucci \etal yield acceptable results, but with visible artifacts. Regarding the synthetic dataset, \method generates the most detailed and photorealistic images. In our results, the eyes of the subjects in the first and second rows show fewer artifacts, and the overall images are considerably cleaner, as can be seen from the tree in the background in the first row.

We provide some video restoration examples in the supplementary material.

\subsection{Ablation Studies}

\begin{table}[t]
    \centering
    \Huge
    \resizebox{\columnwidth}{!}{ 
    \begin{tabular}{lcccc}
    \toprule
    Method & PSNR$\:\uparrow$ & SSIM$\:\uparrow$ & LPIPS$\:\downarrow$ & VMAF$\:\uparrow$\\
    \midrule
    w/o classification & 35.49 & 0.945 & 0.054 & 82.83 \\
    w/o CLIP & 35.42 & 0.945 & 0.053 & 83.37 \\
    \midrule[0.001em]
    w/o MRSFF & 35.37 & 0.944 & 0.055 & 82.94 \\
    w/o att. pooling & \textbf{35.56} & 0.945 & 0.053 & 83.22 \\ \midrule[0.001em]
    $D = 1$ & 35.37 & 0.945 & 0.053 & 83.05 \\
    $D = 3$ & 35.49 & 0.946 & 0.053 & 83.53 \\ \midrule
    \rowcolor{tabhighlight} \textbf{\method} & 35.53 & \textbf{0.946} & \textbf{0.052} & \textbf{83.61} \\
    \bottomrule
    \end{tabular}
    }
    \caption{Ablation studies on the synthetic dataset. Best results are highlighted in bold.}
    \label{tab:ablation}
\end{table}

\paragraph{{Frame Classification and Reference Selection}}
We conduct ablation studies on frame classification and reference selection: 1) \textit{w/o classification}: we select the references from all the frames of the video, without performing frame classification; 2) \textit{w/o CLIP}: we follow the same approach described in \cref{sec:references_classification} but employing the similarity scores and features of BRISQUE \cite{mittal2012no} instead of relying on CLIP and textual prompts. The upper section of \cref{tab:ablation} shows the results. First, we observe that limiting the selection of the references to frames classified as clean improves the results. Differently, degraded frames can be used as references if they are similar enough to the input ones, bringing little improvement to the restoration process. Second, CLIP proves to be more effective than BRISQUE for frame classification. 
See \cref{sec:analysis_frame_classification} for more ablation studies on frame classification.

\paragraph{MRSFF block}
We perform ablation studies on the MRSFF block: 1) \textit{w/o MRSFF}: we substitute the MRSFF blocks with the standard Swin 3D \cite{liu2022video} ones. We make use of spatio-temporal cross-attention by considering the processed features as queries and the reference features as keys and values; 2) \textit{w/o att. pooling}: we substitute attention pooling with a simple average pooling. We report the results in the central section of \cref{tab:ablation}. We notice that the spatio-temporal attention of the Swin 3D blocks performs worse than the spatial-only attention of the MRSFF ones. This is due to the fact that the reference frames are not contiguous in time -- see \cref{sec:references_classification} -- so there is no temporal correlation between them. Moreover, attention pooling proves to be more effective in taking full advantage of the reference frames when compared to simple average pooling. Indeed, it allows us to select the most valuable features with self-attention to combine the information coming from the reference frames into an aggregate representation.

\paragraph{Number of Reference Frames}
We conduct experiments on the number of reference frames $D$, reducing it from 5 to 1 and 3. We expect that a higher number of references will lead to an improvement in the performance, as more information would be available for the processed features to leverage. The lower section of \cref{tab:ablation} shows that our expectation aligns with the experimental results.

%% file: sec/5_conclusion.tex
\section{Conclusion}
In this paper, we present \method, a novel reference-based approach for the restoration of analog videotapes. 
Starting from real-world videos from an archive, we create a synthetic dataset degraded by artifacts typical of analog videotapes. 
We identify the cleanest frames of a video with CLIP by using textual prompts that describe different types of artifact. Then, we exploit those most similar to the degraded input frames as references via the proposed Swin-UNet architecture and the Multi-Reference Spatial Feature Fusion blocks. MRSFF blocks rely on cross-attention and attention pooling to recover the missing details in the input frames. Extensive experiments show the effectiveness of \method compared to several state-of-the-art techniques on both synthetic and real-world datasets. Our results demonstrate the differences between standard and analog video restoration, highlighting the need for approaches specifically designed for this task. In future work, we will develop a learned degradation model, similarly to \cite{bulat2018learn, maeda2020unpaired, luo2022learning}, to efficiently create more accurate synthetic videos.

\paragraph{Acknowledgments}
This work was partially supported by the European Commission under European Horizon 2020 Programme, grant number 101004545 - ReInHerit.

%% file: sec/X_suppl.tex
\clearpage
\maketitlesupplementary

\section{Synthetic Dataset} \label{sec:synthetic_dataset}
\subsection{Dataset Generation}
To generate our synthetic dataset, we start with \textit{4K} videos belonging to the Harmonic dataset \cite{Harmonic-2019}. In particular, we employ videos related to the ``Venice" scenes, which comprise 26,392 frames. We extract the $788 \times 576$ center crop from each frame. Then, we leverage Adobe After Effects \cite{christiansen2013adobe} to add several types of degradation. We aim to make the artifacts in the synthetic frames as similar as possible to those found in real-world videos.

To reach our goal, we first reduce the saturation with the \textit{Hue/Saturation Effect} to make the colors appear duller. Second, we reproduce the tape noise by adding Gaussian noise with the \textit{Noise Effect}. Third, we replicate the tape dropout through an overlay with artifacts typical of VHS tapes blended into the frames in \textit{lighten mode}. Then, we create \textit{six} horizontal grids composed of black and white lines arranged in different ways. A \textit{Wiggle Effect} is applied to the grids to change their vertical position over time without modifying their arrangement. In correspondence with the white lines of the grids, we add horizontal displacement artifacts with the \textit{Displacement Map Effect} to replicate tape mistracking and VHS edge waving. Finally, we use the same approach to reproduce scanline flickering by leveraging the \textit{CC Toner Effect} to blend horizontal cyan, magenta, and green lines into the frames. 

\begin{figure}
  \centering
  \begin{subfigure}{0.495\linewidth}
    \includegraphics[width=\linewidth]{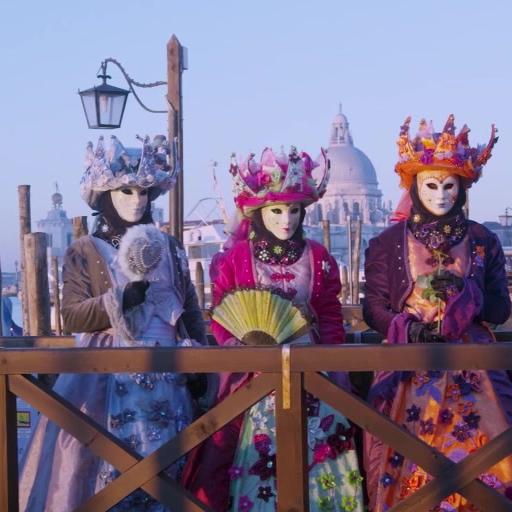}
    \caption{High-quality frame}
    \label{fig:synthetic_frame_a}
  \end{subfigure}
  \hfill
  \begin{subfigure}{0.495\linewidth}
    \includegraphics[width=\linewidth]{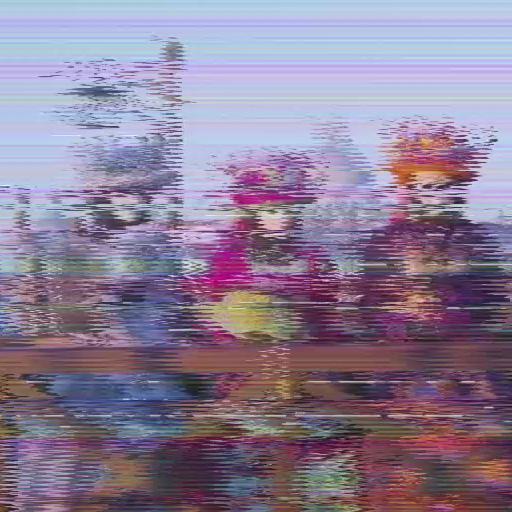}
    \caption{Synthetically degraded frame}
    \label{fig:synthetic_frame_b}
  \end{subfigure}
  \caption{Example of a high-quality frame and the corresponding degraded frame belonging to the synthetic dataset.}
  \label{fig:dataset_frames}
\end{figure}

At this point, our synthetic artifacts look similar to real ones. However, we also need to replicate the time-varying nature of real degradation. Indeed, artifacts in real-world videos change abruptly between consecutive frames and occur at the same time, leading to a disruption of temporal consistency. For this reason, we randomize all the effects we apply to the synthetic videos to make the degradations appear with different intensities, positions, and combinations for each frame. This way, we obtain a dataset of mainly temporally inconsistent videos composed of both almost clean and severely degraded frames, thus resembling real-world videos. \Cref{fig:dataset_frames} shows an example of a high-quality frame and the corresponding synthetically degraded one belonging to our dataset.

\begin{figure}
  \centering
  \begin{subfigure}{0.495\linewidth}
    \centering
    \includegraphics[width=\linewidth]{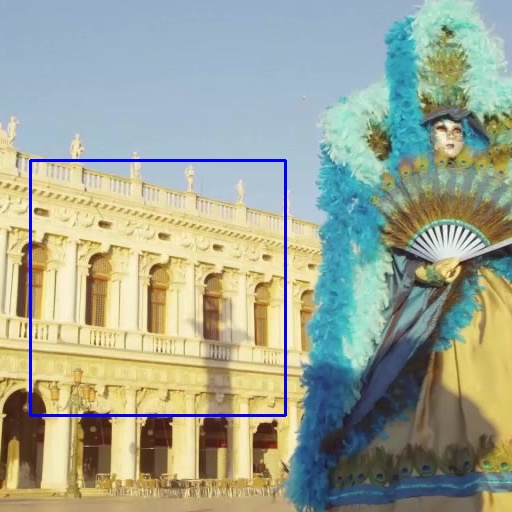}
    \caption{Synthetic clean frame \vspace{5pt}}
    \label{fig:synthetic_clean_frame}
  \end{subfigure}
  \hfill
  \begin{subfigure}{0.495\linewidth}
    \centering
    \includegraphics[width=\linewidth]{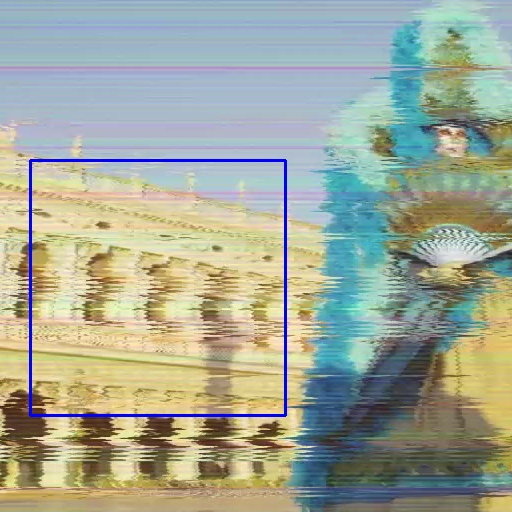}
    \caption{Synthetic degraded frame \vspace{5pt}}
    \label{fig:synthetic_degraded_frame}
  \end{subfigure}
  
   \begin{subfigure}{0.495\linewidth}
    \centering
    \includegraphics[width=\linewidth]{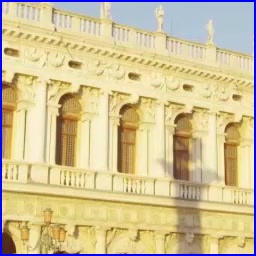}
    \caption{Synthetic clean patch}
    \label{fig:synthetic_clean_crop}
  \end{subfigure}
  \hfill
  \begin{subfigure}{0.495\linewidth}
    \centering
    \includegraphics[width=\linewidth]{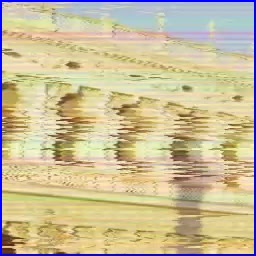}
    \caption{Synthetic degraded patch}
    \label{fig:synthetic_degraded_crop}
   \end{subfigure}
   
   \begin{subfigure}{\linewidth}
    \centering
    \includegraphics[width=\linewidth]{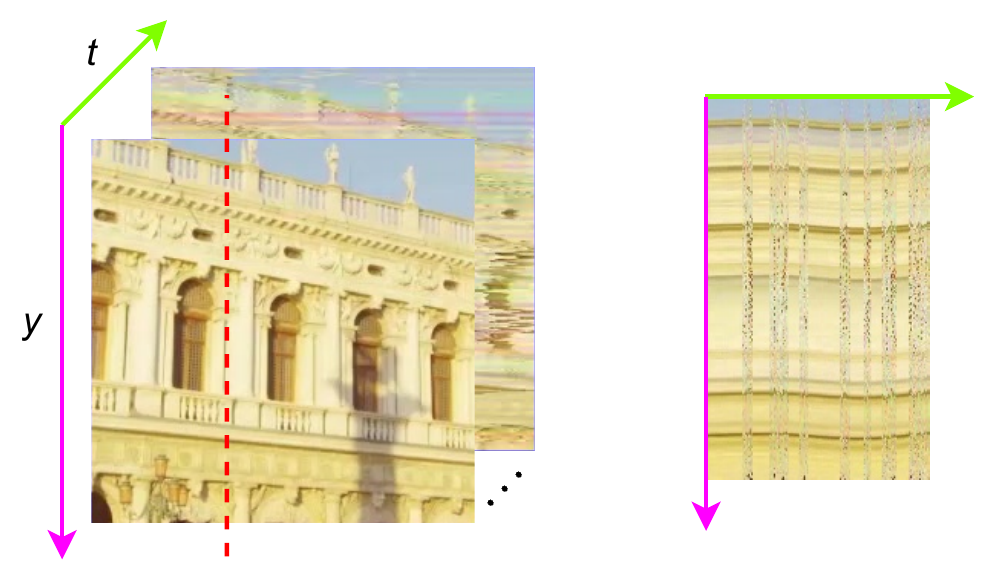}
    \caption{Temporal profile of a column of pixels (red dashed line) belonging to a static synthetic patch. \textit{y} and \textit{t} represent the vertical and temporal axis, respectively.}
    \label{fig:synthetic_input_temporal_profile}
   \end{subfigure}
  
  \caption{Analysis of the degradation and the temporal consistency of a synthetic video.}
  \label{fig:synthetic_comparison}
\end{figure}

\begin{figure}
  \centering
  \begin{subfigure}{0.495\linewidth}
    \centering
    \includegraphics[width=\linewidth]{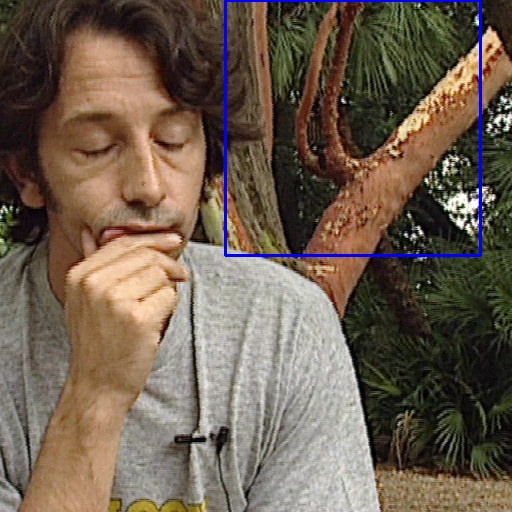}
    \caption{Real-world clean frame \vspace{5pt}}
    \label{fig:original_clean_frame}
  \end{subfigure}
  \hfill
  \begin{subfigure}{0.495\linewidth}
    \centering
    \includegraphics[width=\linewidth]{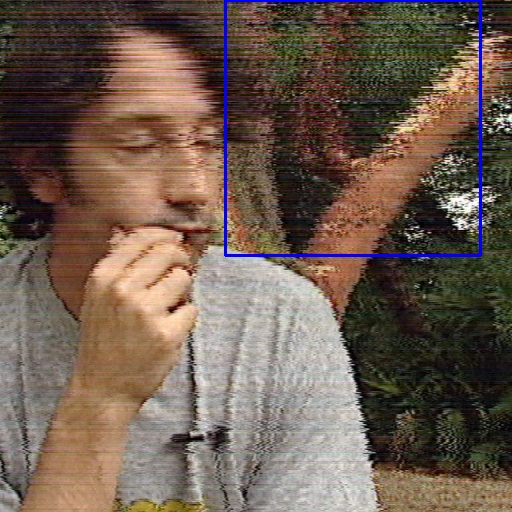}
    \caption{Real-world degraded frame \vspace{5pt}}
    \label{fig:original_degraded_frame}
  \end{subfigure}
  
   \begin{subfigure}{0.495\linewidth}
    \centering
    \includegraphics[width=\linewidth]{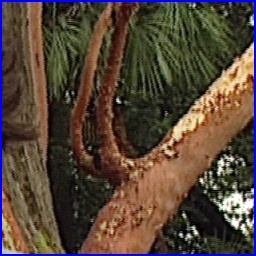}
    \caption{Real-world clean patch}
    \label{fig:original_clean_crop}
  \end{subfigure}
  \hfill
  \begin{subfigure}{0.495\linewidth}
    \centering
    \includegraphics[width=\linewidth]{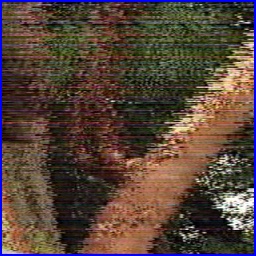}
    \caption{Real-world degraded patch}
    \label{fig:original_degraded_crop}
   \end{subfigure}
   
   \begin{subfigure}{\linewidth}
    \centering
    \includegraphics[width=\linewidth]{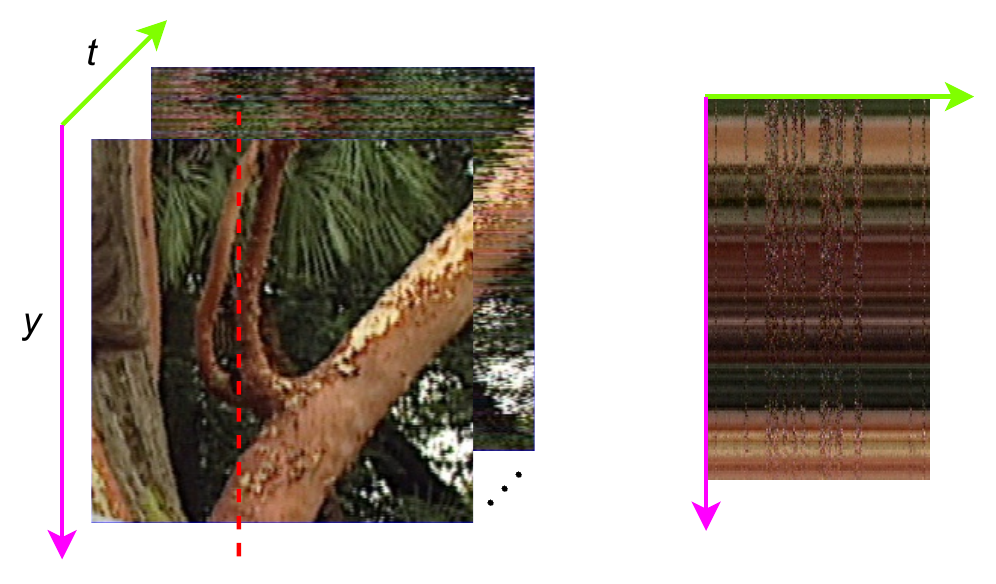}
    \caption{Temporal profile of a column of pixels (red dashed line) belonging to a static real-world patch. \textit{y} and \textit{t} represent the vertical and temporal axis, respectively.}
    \label{fig:original_input_temporal_profile}
   \end{subfigure}
  
  \caption{Analysis of the degradation and the temporal consistency of a real-world video.}
  \label{fig:original_comparison}
\end{figure}

\subsection{Comparison with Real-world Videos}
We provide a qualitative comparison between the synthetic and real-world videos in \cref{fig:synthetic_comparison,fig:original_comparison} to show their similarity. We focus on a static portion of a video, \ie a patch in which both the content and the camera do not move significantly for multiple frames. In this way, we can assess how the appearance of the patch changes between clean and degraded frames and study the temporal consistency. 

\Cref{fig:synthetic_clean_frame,fig:synthetic_degraded_frame} and \Cref{fig:original_clean_frame,fig:original_degraded_frame} illustrate a pair of nearly clean and severely degraded frames belonging to a synthetic and real-world video, respectively. We crop a static patch from each frame and show it in \cref{fig:synthetic_clean_crop,fig:synthetic_degraded_crop} and \cref{fig:original_clean_crop,fig:original_degraded_crop}. We observe that the horizontal colored lines and displacement artifacts in the synthetic frames closely resemble those found in the real-world ones. Moreover, we notice how both the balustrade in \cref{fig:synthetic_degraded_crop} and the branches in \cref{fig:original_degraded_crop} are completely unrecognizable due to the severity of the degradation.

To study the temporal consistency, we select a column of pixels from the static patch of each frame and observe how it varies across time. Without artifacts, we would expect a (nearly) constant temporal profile, since the values of pixels belonging to a still patch do not change as the video progresses. Nevertheless, \cref{fig:synthetic_input_temporal_profile,fig:original_input_temporal_profile} show that in several sequences of frames in both the synthetic and real-world video the temporal consistency is completely lost. Indeed, we observe a smooth temporal transition only for frames in which the considered patches are clean. On the contrary, when the artifacts affect the crops under consideration, the temporal profile contains significant noise. However, the temporal profile also shows how the synthetic video closely resembles the real-world one, proving the quality of our dataset. Moreover, we observe that both the synthetic and real-world videos contain a relatively high quantity of clean frames. Therefore, we always have significantly more than $D$ frames to select the references from. Finally, the temporal profile justifies the main idea underlying our approach: identifying the cleanest frames of each video and exploiting them as references for restoring the sequences of severely degraded frames.

\section{Training Loss} \label{sec:training_loss}
We train our model with a weighted sum of the Charbonnier loss \cite{charbonnier1994two} and a perceptual loss \cite{johnson2016perceptual, ledig2017photo, dosovitskiy2016generating}. Let $I_R$ and $I_{GT}$ be a restored and ground-truth frame, respectively. To make the reconstructed frames faithfully approximate the ground truth ones we employ the Charbonnier loss, defined as:
\begin{equation}
    \mathcal{L}_{char} = \sqrt{\left \| I_{R} - I_{GT} \right \| ^ {2} + \epsilon ^ {2}}
    \label{eq:charbonnier_loss}
\end{equation}
where $\epsilon$ is a constant equal to $10^{-12}$. To improve the perceptual quality and the photorealism of the results, we adopt the perceptual loss \cite{johnson2016perceptual, ledig2017photo, dosovitskiy2016generating}, defined in the VGG-19 \cite{simonyan2014very} feature space. The perceptual loss is formulated as follows:
\begin{equation}
    \mathcal{L}_{perc} = \sum_{l \in L}\frac{1}{C_{l}H_{l}W_{l}} \left\| \Psi_{l}(I_{R}) - \Psi_{l}(I_{GT}) \right\| ^{2}
    \label{eq:perceptual_loss}
\end{equation}
where $C_{l}$, $H_{l}$, $W_{l}$, and $\Psi_{l}$ represent the channel, height, width, and the features from the $l$-th layer of a pre-trained VGG-19 model, respectively, and $L\!=\!\{ \textit{relu2\_2}$, $\textit{relu3\_4}$, $\textit{relu4\_4}$, $\textit{conv5\_4} \}$. Therefore, the overall training loss is:
\begin{equation}
    \mathcal{L} = \lambda_{char} \mathcal{L}_{char} + \lambda_{perc} \mathcal{L}_{perc}
    \label{eq:overall_loss}
\end{equation}
where $\lambda_{char}$ and $\lambda_{perc}$ are the loss weights that we set to 200 and 1, respectively, making their values comparable during training.

\section{Frame Classification}

\subsection{List of Prompts}\label{sec:list_prompts}
As explained in Sec. {\color{red}{3.2}} of the paper, we adopt prompt ensembling \cite{radford2021learning} to improve the frame classification results. The prompts we employ are the following: 1) ``\textit{an image with color artifacts along rows}"; 2) ``\textit{an image with interlacing artifacts}"; 3) ``\textit{an image of a noisy photo}"; 4) ``\textit{an image of a degraded photo}"; 5) ``\textit{a photo with distortions}"; 6) ``\textit{an image of a bad photo}"; 7) ``\textit{a jpeg corrupted image of a photo}"; 8) ``\textit{a pixelated image of a photo}"; 9) ``\textit{a blurry image of a photo}"; 10) ``\textit{a jpeg corrupted photo}"; 11) ``\textit{a pixelated photo}"; 12) ``\textit{a blurry photo}". We crafted prompts $\{1, \ldots, 5 \}$ by converting the artifacts we observed in the real-world dataset into natural language. Prompts $\{6, \ldots, 12 \}$ are more generic and derived from the templates employed for zero-shot classification on the ImageNet dataset by the authors of CLIP \cite{radford2021prompt}. Simply updating the list of prompts makes our approach adapt to different types of degradation, \eg to specialize our method for specific videos or types of medium. We find our list of prompts to be effective, but prompt learning \cite{zhou2022learning} could be considered for future work.

\subsection{Analysis of Frame Classification} \label{sec:analysis_frame_classification}

We evaluate the effectiveness of CLIP in the identification of the cleanest frames of a video by comparing it with several no-reference image quality assessment metrics: BRISQUE \cite{mittal2012no}, NIQE \cite{mittal2012making} and CONTRIQUE \cite{madhusudana2022image}. Since we classify the frames into fairly clean and degraded ones, we can treat it as a binary classification task.

As explained in Sec. {\color{red}3.2}, we compute the histogram of the values of each metric and compute the threshold with Otsu's method for each video of the synthetic test set. Then, we classify as clean every frame with a value of the metric lower than the respective threshold. Additionally, we follow the same procedure for LPIPS, given that it is the most reliable metric to evaluate the perceptual quality of an image. Being a full-reference metric, we can leverage LPIPS only because the synthetic test set has a ground-truth counterpart. We consider the classification performed by LPIPS as the ground truth and the clean frames as positive examples. We measure the performance with standard metrics for binary classification: 1) Accuracy; 2) Precision; 3) Recall 4) F1 score. As can be seen in \cref{tab:reference_selection}, CLIP outperforms all the no-reference metrics. NIQE tends to classify too many degraded frames as clean, achieving a high recall at the cost of lower precision. CLIP proves to be more balanced with the highest accuracy and F1 score. 

\begin{table}[!tbh]
  \centering
  \begin{tabular}{@{}lcccc@{}}
  \toprule
  Metric & Acc$\:\uparrow$ & P$\:\uparrow$ & R$\:\uparrow$ & F1$\:\uparrow$ \\
  \midrule
  BRISQUE \cite{mittal2012no} & 0.745 & 0.657 & 0.801 & 0.710 \\
  NIQE \cite{mittal2012making} & 0.853 & 0.778 & \textbf{0.973} & 0.845 \\
  CONTRIQUE \cite{madhusudana2022image} & 0.862 & 0.823 & 0.931 & 0.862 \\
  CLIP \cite{radford2021learning} & \textbf{0.901} & \textbf{0.881} & 0.889 & \textbf{0.882} \\ 
  \bottomrule
  \end{tabular}
  \caption{Quantitative results for frame classification. $\uparrow$ means that higher values are better. Best results are highlighted in bold.}
  \label{tab:reference_selection}
\end{table}

\begin{figure}
  \centering
  \begin{subfigure}[t]{0.48\linewidth}
    \includegraphics[width=\linewidth]{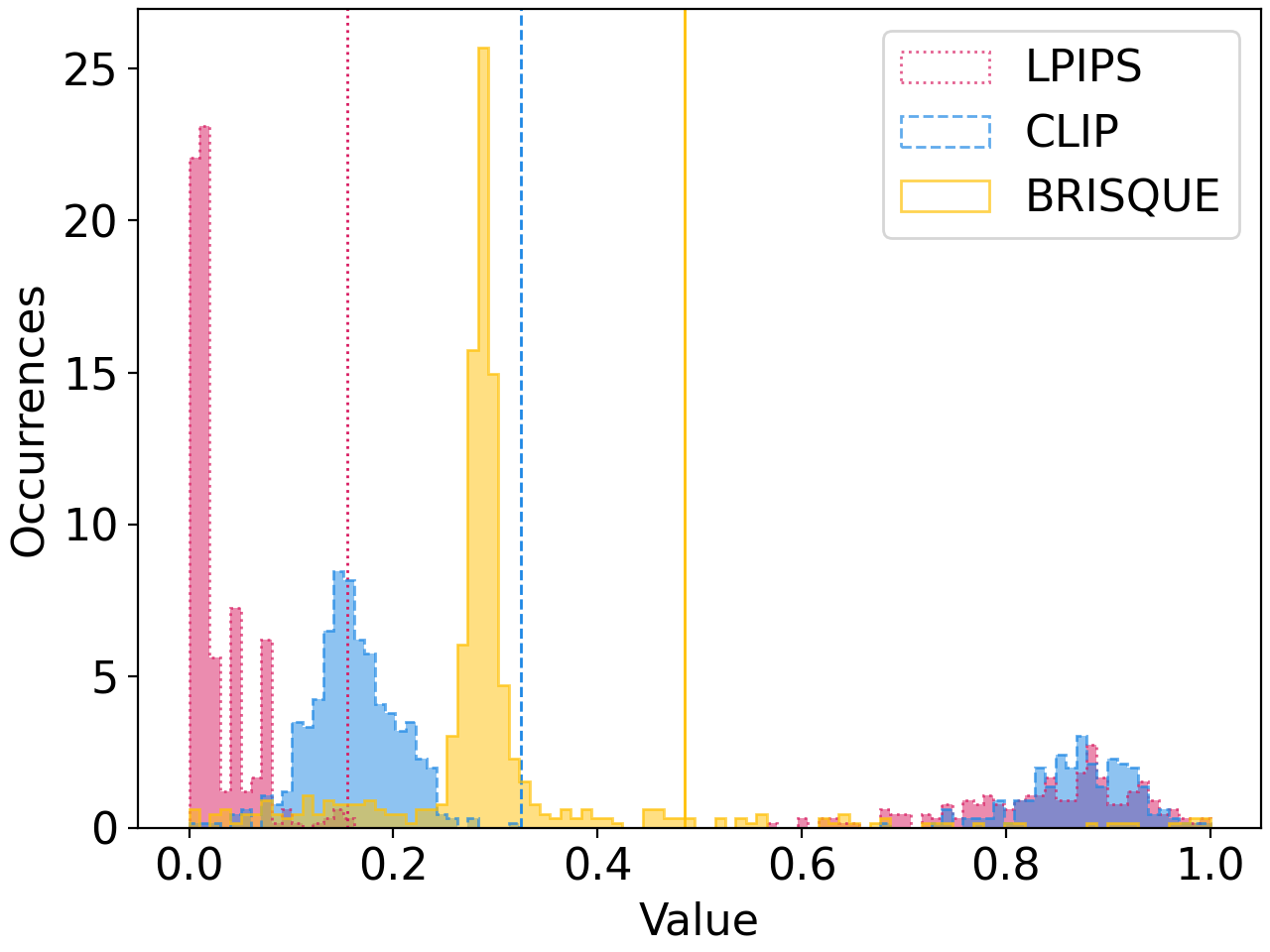}
    \caption{Normalized histograms of the values of the metrics for a given video. The vertical lines represent the threshold values.}
    \label{fig:reference_selection_a}
  \end{subfigure}
  \hfill
  \begin{subfigure}[t]{0.48\linewidth}
    \includegraphics[width=\linewidth]{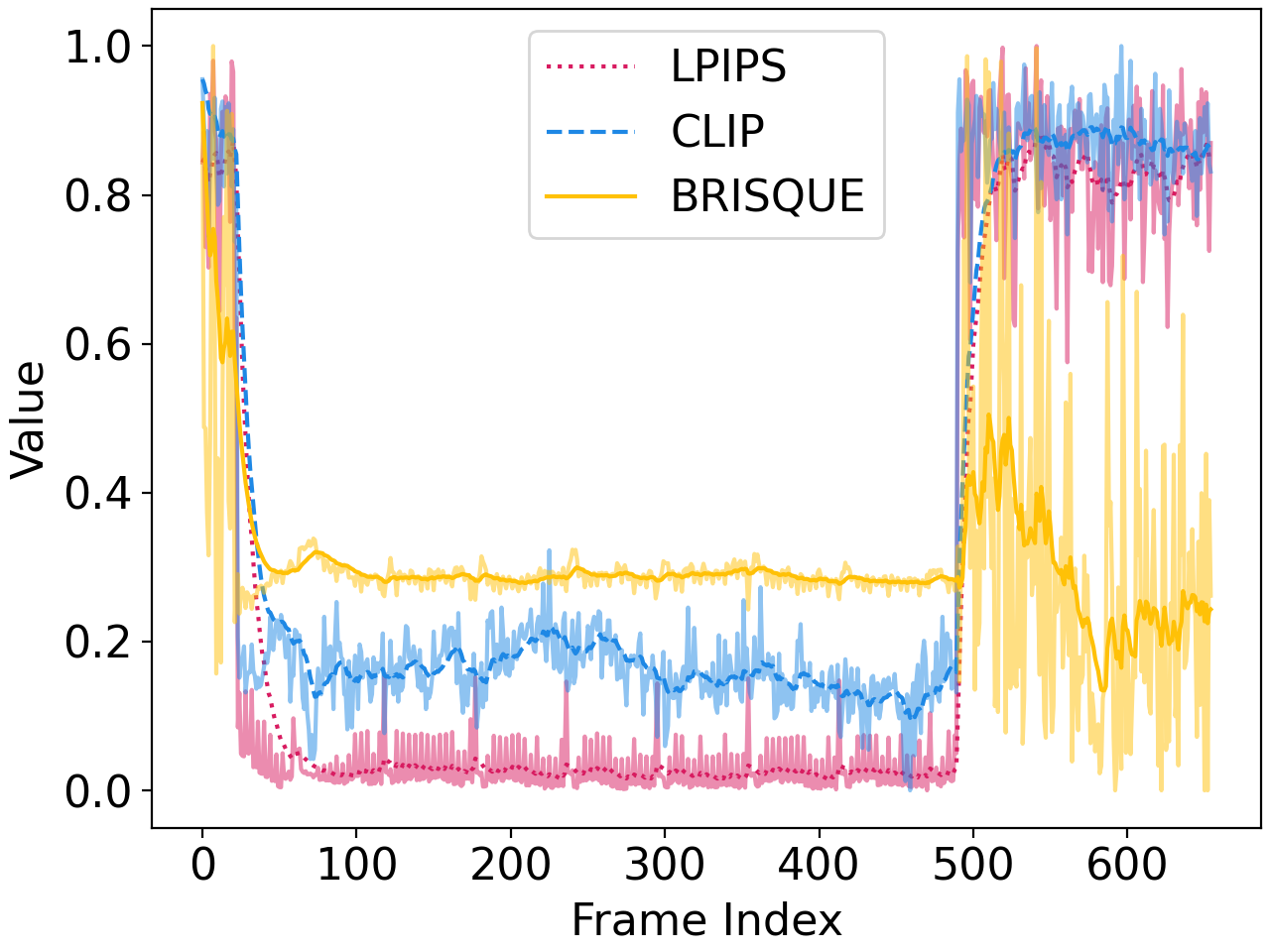}
    \caption{Values of the metrics for a given video stream. The darker lines represent the smoothed values.}
    \label{fig:reference_selection_b}
  \end{subfigure}
  \caption{Visualization of the histograms and of the values of the metrics for a given video. Lower values are better.}
  \label{fig:reference_selection}
\end{figure}

In \cref{fig:reference_selection_a,fig:reference_selection_b} we provide a visualization of the histograms and of the values of the metrics for each frame of a given video, respectively. We exclude NIQE and CONTRIQUE for clearer visualization, but the same considerations made for BRISQUE apply to them. Note that, for all the considered metrics, a lower value corresponds to a higher quality. \Cref{fig:reference_selection_a} shows how the histograms of LPIPS and CLIP are bimodal, while that of BRISQUE is unimodal. This illustrates how CLIP, contrary to BRISQUE, is capable of distinguishing the clean and degraded frames effectively. Looking at the values of LPIPS in \cref{fig:reference_selection_b} shows that the video presents degraded frames at the beginning, a long sequence of clean frames, and then other degraded frames. CLIP manages to capture the profile of the video stream and therefore correctly split the frames into the two classes. On the contrary, BRISQUE values are noisier and some of the degraded frames actually correspond to low values, thus leading to a wrong classification.

We also conduct an ablation study on the use prompt ensembling \cite{radford2021learning} for the frame classification with CLIP. We recall that prompt ensembling improves the robustness of the predictions by averaging the CLIP text features corresponding to multiple prompts. To evaluate the impact of prompt ensembling on the performance, we follow the procedure described in Sec. {\color{red}{3.2}} but rely on a single prompt. In particular, we employ the prompt ``\textit{an image of a degraded photo}" (\ie prompt 4 of the list reported in \cref{sec:list_prompts}). \Cref{tab:reference_selection} reports the results. Using a single prompt corresponds to the highest precision but, as expected \cite{radford2021learning}, prompt ensembling achieves the best overall performance.

\begin{table}[!tbh]
  \centering
  \begin{tabular}{@{}lcccc@{}}
  \toprule
  Metric & Acc$\:\uparrow$ & P$\:\uparrow$ & R$\:\uparrow$ & F1$\:\uparrow$ \\
  \midrule
  Single prompt & 0.888 & \textbf{0.915} & 0.832 & 0.853 \\
  Prompt ensembling & \textbf{0.901} & 0.881 & \textbf{0.889} & \textbf{0.882} \\ 
  \bottomrule
  \end{tabular}
  \caption{Evaluation of the effectiveness of prompt ensembling for the frame classification with CLIP. $\uparrow$ means that higher values are better. Best results are highlighted in bold.}
  \label{tab:reference_selection}
\end{table}

\begin{figure}
  \centering
  \begin{subfigure}[t]{0.495\linewidth}
    \includegraphics[width=\linewidth]{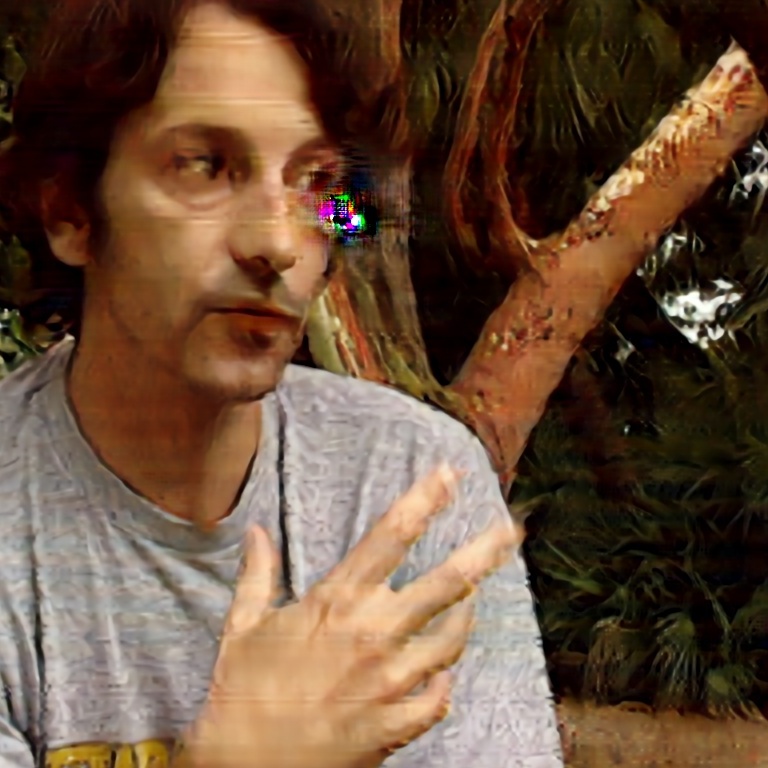}
    \caption{MANA: \textbf{43.77}/\textbf{4.90}/24.42}
  \end{subfigure}
  \hfill
  \begin{subfigure}[t]{0.495\linewidth}
    \includegraphics[width=\linewidth]{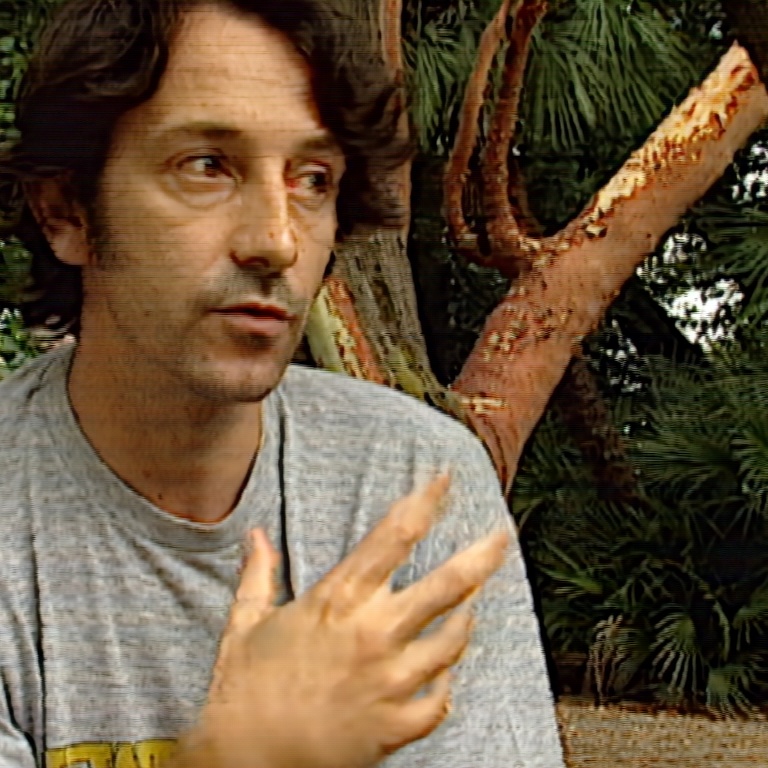}
    \caption{TAPE: 51.49/6.55/\textbf{23.76}}
  \end{subfigure}
  
  \vspace{5pt}
  
  \begin{subfigure}[t]{0.495\linewidth}
    \includegraphics[width=\linewidth]{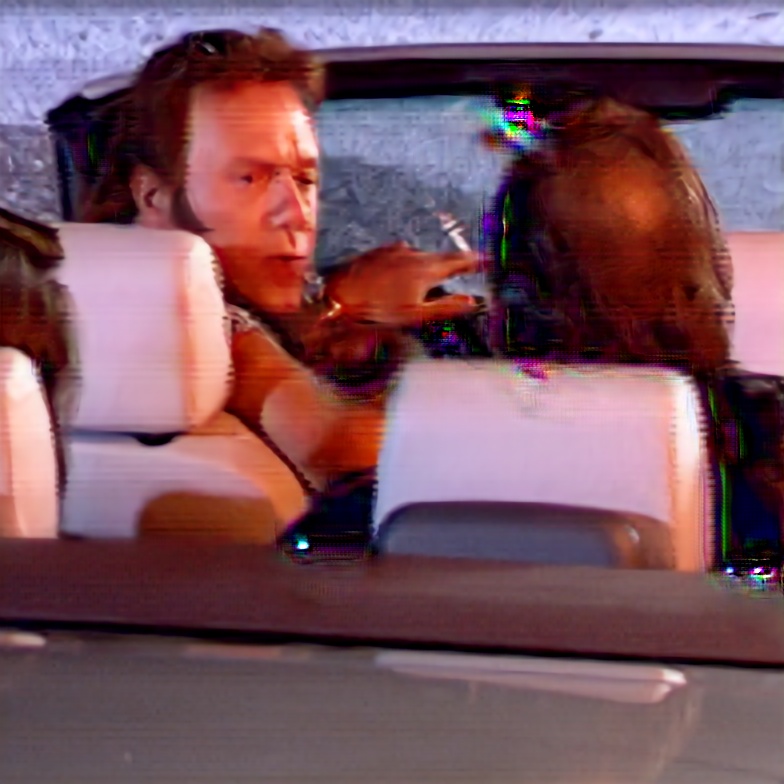}
    \caption{MANA: \textbf{40.70}/\textbf{4.72}/41.86}
  \end{subfigure}
  \hfill
  \begin{subfigure}[t]{0.495\linewidth}
    \includegraphics[width=\linewidth]{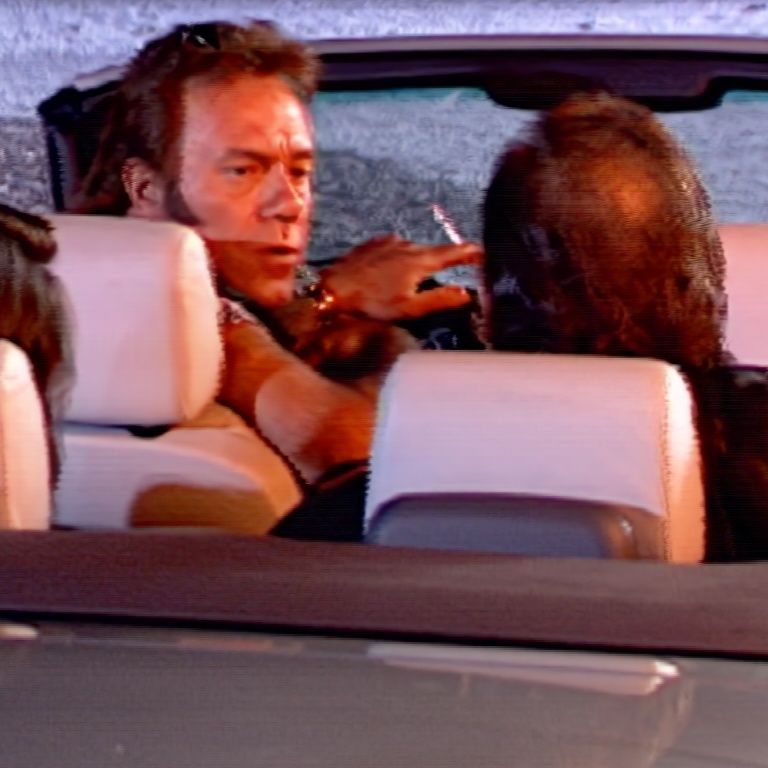}
    \caption{TAPE: 41.27/6.08/\textbf{37.87}}
  \end{subfigure}
  \caption{Comparison between TAPE and MANA \cite{yu2022memory} on frames belonging to real-world videos. The reported values represent BRISQUE$\:\downarrow\,$/NIQE$\:\downarrow\,$/CONTRIQUE$\:\downarrow\,$, respectively, where $\downarrow$ means that lower values are better. Best results for each pair of images are highlighted in bold.}
  \label{fig:quantitative_results_analysis}
\end{figure}

\section{Analysis of Quantitative Results} \label{sec:analysis_quantitative_results}
In Sec. {\color{red}4.4} of the paper, we report the quantitative results for the real-world dataset. Our approach achieves the best results for CONTRIQUE, while MANA \cite{yu2022memory} is better for BRISQUE and NIQE. However, the qualitative results show that MANA adds high-frequency artifacts to restored frames. We argue that these artifacts deceive BRISQUE and NIQE, which mistake them for high-frequency details that are distinctive of high-quality images \cite{seidenari2022language,agnolucci2023perceptual}. In \cref{fig:quantitative_results_analysis} we show two real-world examples supporting our argument. We report images obtained with TAPE and MANA and the corresponding BRISQUE, NIQE and CONTRIQUE values. Our approach generates results that are clearly more satisfying and photorealistic and achieves the best values for CONTRIQUE. However, MANA obtains the highest BRISQUE and NIQE values, despite the presence of highly visible artifacts. We suppose that these artifacts are caused by the use of the memory bank learned during the training with our synthetic dataset which makes MANA fail to generalize to a different dataset. The lack of these artifacts in the results of the synthetic test set supports our hypothesis.